%% file: main_arxiv.tex
\documentclass{l4dc2027}

\jmlrvolume{}
\jmlryear{}
\jmlrpages{}
\jmlrproceedings{}{}

\input{math_commands.tex}

\input{math_definitions.tex}

\usepackage{times}
\usepackage{paralist}
\usepackage{booktabs}
\usepackage{float}
\usepackage{caption} % better captions
\usepackage{wrapfig}
\usepackage{xcolor}
\usepackage[most]{tcolorbox}
\usepackage{xurl}

\newcommand{\RomanNum}[1]{\MakeUppercase{\romannumeral #1}}
\newcommand{\solver}{$\phi$-PICT}

\definecolor{authA}{HTML}{0072B2}
\definecolor{authB}{HTML}{D55E00}
\definecolor{authC}{HTML}{009E73}
\definecolor{authD}{HTML}{CC79A7}
\definecolor{authE}{HTML}{E69F00}

\definecolor{BrightLightIndigo}{HTML}{E8EAF6}
\definecolor{BrightIndigo}{HTML}{3F51B5}

\title[Sample-Efficient Control of Magnetoconvective Flows via Differentiable Simulation]{Sample-Efficient Control of Magnetoconvective Flows via Differentiable Simulation}

\author{%
 \Name{Jannis Becktepe} \Email{jannis.becktepe@tu-dortmund.de}\\
 \addr TU Dortmund University \& Lamarr Institute for Machine Learning and Artificial Intelligence
 \AND
 \Name{Dmitry Krasnov} \Email{dmitry.krasnov@tu-ilmenau.de}\\
 \addr Institute of Thermodynamics and Fluid Mechanics, TU Ilmenau
 \AND
 \Name{Nils Thuerey} \Email{nils.thuerey@tum.de}\\
 \addr Physics-based Simulation Group, TU Munich \& Munich Center for Machine Learning
 \AND
 \Name{J{\"o}rg Schumacher} \Email{joerg.schumacher@tu-ilmenau.de}\\
 \addr Institute of Thermodynamics and Fluid Mechanics, TU Ilmenau
 \AND
 \Name{Sebastian Peitz} \Email{sebastian.peitz@uni-paderborn.de}\\
 \addr Scientific Machine Learning Group, Paderborn University%
}

\begin{document}

\maketitle

\begin{abstract}%
Magnetohydrodynamic~(MHD) fluid flows are central to demanding engineering applications, most prominently fusion blanket cooling, where flow control is a promising tool to optimize key properties such as heat transfer.
While active flow control~(AFC) has recently advanced through machine learning, its application to MHD flows remains largely unexplored.
In this work, we present \solver{}, a novel differentiable GPU-accelerated MHD solver targeted at AFC.
We validate \solver{} against MHD benchmarks and demonstrate its high computational efficiency.
Integrated into FluidGym, \solver{} enables sample-efficient gradient-based algorithms to improve heat transfer by $>25\%$ and allows optimizers to achieve $10^3\times$ higher accuracy in inverse design than black-box methods.
\solver{} is publicly available at \url{https://github.com/upb-sciml/phiPICT}.
\end{abstract}


\section{Introduction}
Control of large-scale physical systems governed by high-dimensional, non-linear partial differential equations~(PDEs) remains a fundamental challenge at the intersection of machine learning, physics, and control theory~\citep{brunton2020}.
Among such systems, magnetohydrodynamics~(MHD) introduces distinct control challenges driven by the interaction between an electrically conducting fluid and external magnetic field.
In thermal flows, this coupling leads to so-called \emph{magnetoconvection}, which fundamentally alters the flow structures and heat transfer. 
Liquid metal magnetoconvection is key to many engineering applications, such as cooling in nuclear-power facilities~\citep{rachkov2018a} or fusion reactor blankets~\citep{aiello2014, zmitko2016}.

In fusion blanket duct flows, \citet{zikanov2016} showed that strong temperature fluctuations can be explained by wall jets destabilizing into large quasi-2D~(Q2D) vortices that induce strong mixing.
Subsequent works have explored passive strategies to mitigate these thermal stresses, including asymmetric heat loads~\citep{belyaev2020} and modified inlet  geometries~\citep{zikanov2019}.
However, passive strategies cannot adapt dynamically to non-stationary, evolving turbulent structures.
Consequently, closed-loop active flow control~(AFC) emerges as a promising candidate for real-time flow stabilization and heat transfer improvement in magnetoconvective fluid dynamics.

\noindent
Data-driven approaches, particularly reinforcement learning~(RL), have emerged as a promising technique for AFC and have achieved notable successes in drag reduction, heat transfer enhancement, and flow separation~\citep{rabault2019, beintema2020, vignon2023a, guastoni2023, montala2026, Straat2026robustness}.
Concurrently, differentiable simulators enable direct backpropagation through physical dynamics and can substantially improve sample-efficiency over traditional RL~\citep{becktepe2026,tomasetto2026}.
Despite these advances, the application of RL and gradient-based control to liquid metal flows in closed channels or ducts remains largely unexplored, primarily due to the lack of differentiable, RL-compatible MHD simulators.

To bridge this gap, we introduce \solver{}, a GPU-accelerated, differentiable MHD solver and integrate it into the open-source FluidGym framework.
In summary, our main contributions are
\begin{inparaenum}[(1)]
    \item \textbf{$\boldsymbol{\phi}$-PICT Solver:} A differentiable, GPU-accelerated MHD solver for low magnetic Reynolds numbers that enables end-to-end backpropagation through coupled MHD and thermal dynamics and is validated against canonical MHD and convection direct numerical simulation benchmarks;
    \item \textbf{FluidGym Integration:} A modular 3D liquid metal Rayleigh-Bénard convection control environment in FluidGym for black-box RL and gradient-based policy learning; and
    \item \textbf{Empirical Evaluation:} Demonstrations showing that gradient-based optimization via \solver{} is substantially more sample-efficient across both domains, outperforming black-box RL to increase heat transfer by $>25\%$ and achieving 3 orders of magnitude higher accuracy in inverse wall-defect detection\footnote{Run scripts: \url{https://github.com/upb-sciml/fluidgym/tree/mhd_control/runscripts}}.
\end{inparaenum}

\section{Related Work}
\label{sec:related_work}

\paragraph{Numerical Solvers for MHD Flows.}
Simulating liquid metal flows in the presence of strong magnetic fields in fusion reactors typically relies on the inductionless, quasistatic approximation with magnetic Reynolds numbers $\mathrm{Rm} \ll 1$~\citep{davidson2001}.
Common numerical software includes \texttt{epotFOAM}~\citep{tassone2016}, based on \texttt{openFOAM}~\citep{weller1998a} and specialized algorithms like Tensor-Product Thomas solvers~\citep{krasnov2023}.

\paragraph{Differentiable Solvers.}
Differentiable fluid solvers~\citep{holl2024,bezgin2025,franz2026} have become a promising tool for general PDE control~\citep{holl2020a} and gradient-based policy learning~\citep{becktepe2026,zanotta2026,tomasetto2026}. 
Existing differentiable MHD frameworks focus on high-$\mathrm{Rm}$ astrophysics in geometries with simpler boundary conditions than those at solid walls~\citep{storcks2026}. 
To the best of our knowledge, no differentiable solver exists for low-$\mathrm{Rm}$ magnetoconvective flows in closed geometries.

\paragraph{Reinforcement Learning for Active Flow Control.}
Recently, deep RL has shown great success in closed-loop flow control.
Studies range from aerodynamic drag reduction~\citep{rabault2019,garcia2025b,montala2026} to convective heat transfer enhancement~\citep{beintema2020}.
For spatially distributed control, multi-agent reinforcement learning~\citep[MARL; see][for an introduction]{albrecht2024} has shown promising results in terms of generalization and sample efficiency~\citep{guastoni2023,vignon2023a,vasanth2024,markmann2025}.
In nuclear fusion research, RL has been used for magnetic control of plasma in Tokamak reactors~\citep{degrave2022,mouchamps2026a}.
However, RL for AFC has not yet been applied to liquid metal flows present, for instance, in cooling blankets.
We address this gap by introducing a differentiable MHD control benchmark integrated into FluidGym~\citep{becktepe2026}, an open-source plug-and-play benchmark for RL and gradient-based methods for AFC.

\section{\texorpdfstring{$\bm{\phi}$}{\phi}-PICT: Differentiable Simulation of Magnetoconvective Fluid Flows}

\begin{wrapfigure}[9]{r}{0.49\textwidth}
    \vspace{-0.17em}
    \centering
    \includegraphics[width=\linewidth]{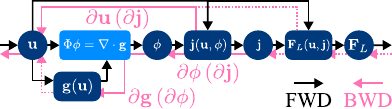}
    \caption{Computation of $\lorentz$ given $\vel$ in \solver{}. Solid and dotted lines denote explicit analytical routines and \texttt{autograd}, respectively.}
    \label{fig:solver_steps}
\end{wrapfigure}

Our differentiable solver \solver{} for MHD fluid dynamics builds on PICT~\citep{franz2026}, which provides a differentiable PISO loop, basic boundary conditions, and the adjoint mechanism for the standard velocity and pressure linear solves.
To adapt this to MHD, \solver{} implements $\epot$-PISO~\citep[][see Appendix~\ref{sec:app:solver} for details]{masdelesvalls2011}.
Specifically, it adds
\begin{inparaenum}[(1)]
    \item the electric potential Poisson solver,
    \item a conservative current density formulation, 
    \item the Lorentz force coupling, and crucially,
    \item the custom gradient functions required to backpropagate through the new PISO steps related to the electric potential, current density, and Lorentz force (see Figure~\ref{fig:solver_steps}).
\end{inparaenum}

\paragraph{Governing Equations.}
Following previous works~\citep{belyaev2020}, we consider the so-called \emph{inductionless} or \emph{quasi-static} approximation. 
Here, the magnetic field is considered static and not affected by the fluid \citep[see][for details]{davidson2001}.
Under this formulation, the incompressible momentum and continuity equations are
\begin{align}
    \frac{\partial \vel}{\partial t} + (\vel \cdot \nabla)\vel = - \nabla p + \frac{1}{\reyn} \nabla^2 \vel + \lorentz + \buoyancy \quad \text{and} \quad \nabla \cdot \vel = 0\,. \label{eq:momentum_continuity}\
\end{align}
The thermal transport and the buoyancy force based on the Boussinesq approximation are given by
\begin{align}
\frac{\partial \temp}{\partial t} + \vel \cdot \nabla \temp = \frac{1}{\reyn\pr} \nabla^2 \temp \quad \text{with} \quad \buoyancy = -\frac{\mathrm{Gr}}{\reyn^2}\egrav (\temp - \temp_0)\,. \label{eq:heat_transfer_buoyancy_force}
\end{align}
Finally, the current density, charge conservation, and Lorentz force are defined as
\begin{align}
\currDens = - \nabla \epot + \vel \times \eb \quad \text{and} \quad \nabla \cdot \currDens = 0 \quad \text{with} \quad \lorentz = \frac{\ha^2}{\reyn} \currDens \times \eb\,. \label{eq:current_density_charge_conservaton_lorentz_force}
\end{align}
The non-dimensional parameters of the coupled magnetohydrodynamic and convective heat transfer system are the Reynolds number $\reyn = U H / \nu$, Hartmann number $\ha = B_0 H \sqrt{\sigma / (\rho \nu)}$, Prandtl number $\pr = \nu / \alpha$, Grashof number $\gr = g \beta \Delta \temp H^3 / \nu^2$, and Stuart number $\stuart = \ha^2 / \reyn$.
Here, $U$ is the characteristic velocity, $H$ is the characteristic length scale, $\nu$ is the kinematic viscosity, $B_0$ is the magnetic field magnitude, $\sigma$ is the electrical conductivity, $\rho$ is the fluid density, $\alpha$ is the thermal diffusivity, $g$ is the gravitational acceleration, $\beta$ is the thermal expansion coefficient, $\temp_0$ is the reference temperature, and $\Delta \temp$ is the characteristic temperature difference.

\paragraph{Forward Simulation.}
An overview of the forward simulation steps computing $\lorentz$ from $\vel$ is shown in Figure~\ref{fig:solver_steps}.
From the charge conservation in Equation~\ref{eq:current_density_charge_conservaton_lorentz_force}, we can derive the Poisson equation for the electric potential $\epot$: via $\nabla \cdot \currDens = 0 \Leftrightarrow \nabla^2 \epot = \nabla \cdot (\vel \times \eb)$.
Following \citet{masdelesvalls2011,tassone2016}, we apply the conservative Lorentz force scheme proposed by \citet{ni2007}.

\paragraph{Gradient Computation.}
To enable end-to-end backpropagation through \solver{}, we derive analytical gradients alongside the forward simulation steps (see Figure~\ref{fig:solver_steps}).
Following \citet{franz2026}, $\partial y$ denotes the gradient of the loss w.r.t.\ $y$, and the vector-Jacobian product $\gradfn{x}{y} = (\partial / \partial x)^\top \partial y$ yields the contribution to $\partial x$ via $y$.
Backpropagation starts from $\partial \lorentz$ and reaches $\vel$ through $\uxB$ along two paths: directly via the face current fluxes, and indirectly via the electric potential $\epot$.
Analytical gradients are derived for two core operations:
\begin{inparaenum}[(i)]
    \item the conservative face-based current density construction, which couples adjacent cells via face flux sensitivities to yield $\gradfn{\epot}{\currDens}$ and $\gradfn{(\uxB)}{\currDens}$, and
    \item the linear potential solve.
\end{inparaenum}
For the discrete linear system of the electric potential $\EpotMat \epot = \epotRHS$ with discrete potential matrix $\EpotMat$ and $\epotRHS = \nabla \cdot \epotRHSvec = \nabla \cdot (\uxB)$, we solve the adjoint system $\EpotMat^T \partial \epotRHS = \partial \epot$ to obtain $\partial g$, which gives the gradient function $\gradfn{\epotRHS}{\epot}$.
In contrast to naive unrolling of Conjugate Gradient~(CG), solving the adjoint allows $\mathcal{O}(1)$ memory scaling w.r.t.\ linear solve iterations.
Standard cell operations, such as the cross products producing $\partial \currDens$ and finally $\partial \vel = \magn \times \partial(\uxB)$, are handled by PyTorch \texttt{autograd} (dotted lines in Figure~\ref{fig:solver_steps}).
Although $\magn$ is assumed constant in this work, \solver{} also enables gradient backpropagation with respect to $\magn$.
Detailed gradient derivations are stated in Appendix~\ref{sec:app:solver:gradients}.

\paragraph{Solver Validation.}
In the following, we describe how we thoroughly evaluate our MHD solver.
First, we run Hartmann, Shercliff, and Hunt flows and match the analytical velocity profiles. 
A grid refinement study shows how the relative errors decrease with higher resolutions.
Furthermore, we compare our results to the jet detachments in a duct flow discussed by \citet{arlt2018}.
To further validate our solver for \emph{magnetoconvective} flows, we run validation cases spanning various Hartmann numbers and wall conduction ratios that match results previously published by \citet[][see Tables~\ref{tab:app:val_temp1}~and~\ref{tab:app:val_temp2}]{krasnov2023}.
To validate the correctness and informativeness of simulation gradients, we run the PyTorch \texttt{gradcheck} tool as well as an optimization task where we learn the Stuart number of a reference simulation via backpropagation.
Together, these results show the correctness of our proposed solver in both forward and backward passes (see Appendix~\ref{sec:app:validation} for detailed results).

\paragraph{Further Improvements.}
\begin{wrapfigure}{r}{0.49\textwidth}
    \centering
    \includegraphics[width=\linewidth]{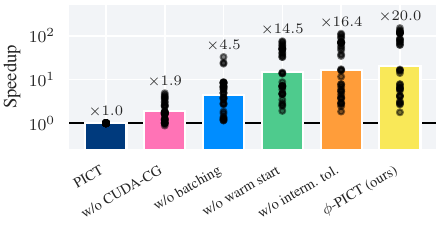}
    \caption{Geometric mean runtime improvements of \solver{} over PICT across all non-MHD FluidGym environments. Dots indicate speedup factors for individual environments.}
    \label{fig:solver_ablation_timing}
\end{wrapfigure}

Due to the large computational cost associated with fluid simulations, we incorporate a series of improvements into \solver{} to enhance its efficiency and consequently the downstream performance of FluidGym for (magnetoconvective) environments.
To fully utilize the GPU, we introduce batched simulations of over $250$ parallel FluidGym environments, yielding substantial speedups over FluidGym based on PICT.
To further reduce simulation runtime at scale, we introduce three core optimizations to the CG solver in \solver{}:
\begin{inparaenum}[(i)]
    \item transferring the whole CG solve execution to the GPU without intermediate CPU communication,
    \item warm-starting the pressure solve from the previous time step, and
    \item adopting adaptable solver tolerances during intermediate iterations, following \texttt{OpenFOAM}.
\end{inparaenum}
Consequently, \solver{} achieves a substantial $20\times$ average speedup over the baseline PICT solver in FluidGym.
Notably, for some environments, speedups of up to approx. $150\times$ are achieved (see Table~\ref{tab:app:solver:solver_benchmark} in Appendix~\ref{sec:app:solver:further_improvements}).
Specifically for MHD, we additionally leverage preconditioning of the electric potential Poisson linear system through the algebraic multigrid~\citep[AMG;][]{ruge1987} method via \texttt{PyAMG}~\citep{bell2023}.

\section{Learning Closed-loop Control: Liquid Metal Rayleigh-Bénard Convection}

\subsection{Experimental Setup}
\begin{wrapfigure}{r}{0.49\textwidth}
    \centering
    \includegraphics[width=\linewidth]{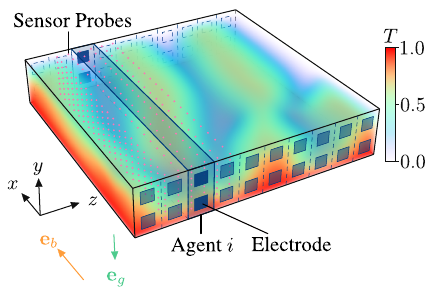}
    \caption{Uncontrolled MHD RBC setup with electrodes at the Hartmann walls at $\rayleigh = 50\chandr$.}
    \label{fig:control_setup}
\end{wrapfigure}

In the following, we present a novel control setup for 3D liquid metal Rayleigh-Bénard convection~(RBC) in a closed box of size $L_x \times L_y \times L_z = 5 \times 1 \times 5$ following \citet{yanagisawa2013} that is heated at the bottom and cooled from the top.
The setup is visualized in Figure~\ref{fig:control_setup}.
Motivated by the idea that quasi-two-dimensional convection rolls have desirable properties in fusion blankets, the control environment aims to stabilize these rolls.
The stability of the rolls depends on the ratio of the Rayleigh number $\rayleigh$ to the Chandrasekhar number $\chandr = \ha^2$.
When fixing $\chandr$ and increasing $\rayleigh$, the flow exhibits decreasing magnetic damping and higher instability, which increases control difficulty.
As studied by \citet{yanagisawa2013}, there are distinct flow regimes one could consider regarding $\rayleigh$ in terms of $\chandr$.
Particularly, these can be characterized by
\begin{inparaenum}[(i)]
    \item a transition from five to four rolls between $10\chandr$ and $30\chandr$; and
    \item a four-roll oscillation and a transition from four to three rolls between $30\chandr$ and $50\chandr$.
    % \item a fluctuating cell without rolls for $\rayleigh > 100\chandr$.
\end{inparaenum}
Based on these findings, we consider $\rayleigh = 20\chandr$ and  $\rayleigh = 50\chandr$.

\paragraph{Numerical Setup.}
We simulate the flow with \solver{}, solving Equations~\ref{eq:momentum_continuity}--\ref{eq:current_density_charge_conservaton_lorentz_force} with a uniform horizontal magnetic field $\eb = \bm{e}_x$ and gravity $\egrav = -\bm{e}_y$.
We use $\pr = 0.025$ and $\chandr = 10^3$.
All walls are no-slip and electrically insulating.
The bottom and top plates are isothermal at $\temp = 1$ and $\temp = 0$, with $\temp_0 = 0.5$, and the side walls are adiabatic.
The magnetic field damps velocity gradients along $\eb$, so the rolls align with $x$ and are stacked along $z$, and the walls at $x = 0$ and $x = L_x$ are the Hartmann walls, i.e., the walls perpendicular to the magnetic field.
The grid has $160 \times 32 \times 160$ cells, with $\tanh$  refinement towards all walls.
Along $x$, the grading resolves the Hartmann layers of thickness $\delta \sim H/\ha$ with at least three cells.
The time step adapts to a CFL number of $0.8$ with control step length $T_\mathrm{control} = 1.5$ free-fall time units.
Discrete control steps are denoted by index $k$.
A set of 30 initial states for training, validation, and testing is obtained by evolving the conduction profile with small temperature perturbations for 30 free-fall times.
At each reset, we randomize the roll configuration by applying a random combination of mirroring, small temperature noise, and a short random warm-up similar to the procedures in FluidGym~\citep{becktepe2026}.

\paragraph{Actuation, Observations, Reward.}
Following \citet{becktepe2026}, we formalize the control task as a partially observable stochastic game~(POSG) for multi-agent reinforcement learning~(MARL) based on \citet{albrecht2024}.
In particular, we consider $M$ individual agents acting in the fluid domain.
While all agents share the same policy, each agent with index $i \in [0, \dots, M-1]$ acts locally via $a_i$  and receives an individual reward $r_i$ and observation $o_i$.
Here, each agent corresponds to one of $M = 10$ electrode stations distributed along $z$ on both Hartmann walls.
Each station holds two electrodes along $y$, so agent $i$ sets two normalized currents $a^i \in [-1, 1]^2$.
Opposing electrodes on both walls carry the same current, keeping the forcing symmetric about $x = L_x/2$.
Since the box is otherwise insulating, the joint action is projected onto zero net current before being applied.
A zero action thus recovers the uncontrolled flow.
Each agent observes 960 velocity and temperature sensors on the full $x$–$y$ grid, within a local z-window that covers its own station and its neighbours' actuators, padded beyond the side walls with zero velocity and the conduction-profile temperature. 
The global reward at control step $k$ balances heat transfer enhancement against electrical power consumption:
\begin{equation}
    r_k = \frac{\nusselt_k}{\nusselt_\mathrm{ref}} - 1 - \lambda\,\frac{P_{\mathrm{el},k}}{P_\mathrm{ref}},
    \nonumber
\end{equation}
where both terms are averaged over the inner solver steps of step $k$.
Here, $\nusselt = 1 + \sqrt{\mathrm{Ra}\,\mathrm{Pr}}\,\langle u_y\,(\temp - \temp_0) \rangle_V$ is the volume-averaged Nusselt number.
Quantities indexed with $\mathrm{ref}$ denote the uncontrolled baseline.
The injected electrical power $P_\mathrm{el} = N \int_{S_\mathrm{e}} \phi_\mathrm{w}\, j_\mathrm{in} \,\mathrm{d}S$ over electrode surface $S_\mathrm{e}$ is normalized by the thermal driving power $P_\mathrm{ref} = \int_V u_y\,(\temp - \temp_0)\,\mathrm{d}V$.
Following \citet{vignon2023a}, individual rewards $r^i_k = \beta\,(\nusselt^i_k/\nusselt_\mathrm{ref} - 1) + (1 - \beta)\,r_k$ blend local and global quantities.

\paragraph{Control Algorithms.}
As a black-box RL algorithm we consider Soft Actor-Critic~\citep[SAC; ][]{haarnoja2018a} implemented in the Stable-Baselines3~\citep[SB3; ][]{raffin2021} framework using its default hyperparameters.
To leverage the differentiable simulation, we train policies using backpropagation through time~\citep[BPTT/APG;][]{werbos1990,freeman2021} and Short-Horizon Actor-Critic~\citep[SHAC;][]{xu2021}.
In both approaches, policy parameters are updated by backpropagating the gradient of the discounted return over a horizon of $10$ control steps through the physics simulation.
Notably, the naive BPTT algorithm does not leverage explicit exploration, value function estimation, or experience replay.
All algorithms are trained across $5$ random seeds on the FluidGym \texttt{train} split.
After training, policies are evaluated deterministically on $10$ \texttt{test} episodes.
Detailed hyperparameters are stated in Appendix~\ref{sec:app:experiments:rl_hps}.

\subsection{Results}
\begin{figure}[tpb]
    \centering
    \includegraphics[width=\textwidth]{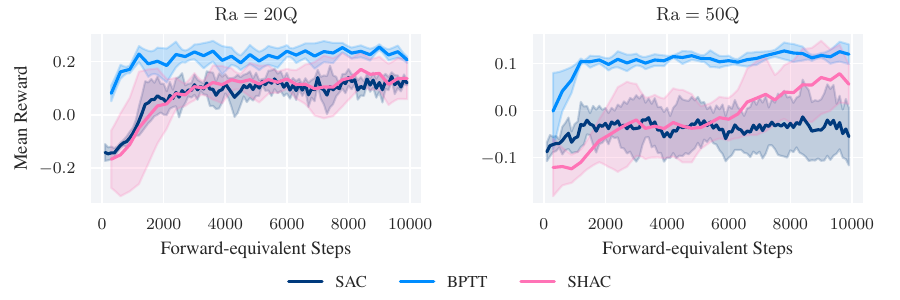}
    \caption{Mean training reward per step over forward-equivalent control steps for BPTT, SAC, and SHAC (accounting for the $\sim 3\times$ step cost of backpropagation and activation checkpointing). Solid lines and shaded areas represent mean $\pm$ standard deviation across seeds, respectively.}
    \label{fig:RBCMHD3D_reward_combined}
\end{figure}
To account for the $3 \times$ computational overhead of activation checkpointing and backward passes for differentiable methods, we measure training progress in forward-equivalent simulation steps.
Figure~\ref{fig:RBCMHD3D_reward_combined} shows the mean training reward trajectories. %for BPTT, SAC, and SHAC across both regimes.
Notably, BPTT shows the fastest convergence and reaches its final reward after $\approx 1\,000$ forward-equivalent steps, corresponding to $\approx 333$ environment steps.
SHAC and SAC perform similarly at $\rayleigh = 20\chandr$, while SHAC slightly outperforms SAC at $\rayleigh = 50\chandr$.
However, neither SAC nor SHAC reaches the reward achieved by BPTT throughout $10$k forward-equivalent training steps.
Notably, despite lacking SHAC's stabilization mechanisms, BPTT proves more effective here, demonstrating the sample-efficiency enabled by \solver{}.
\begin{figure}[tpb]
    \centering
    \includegraphics[width=\textwidth]{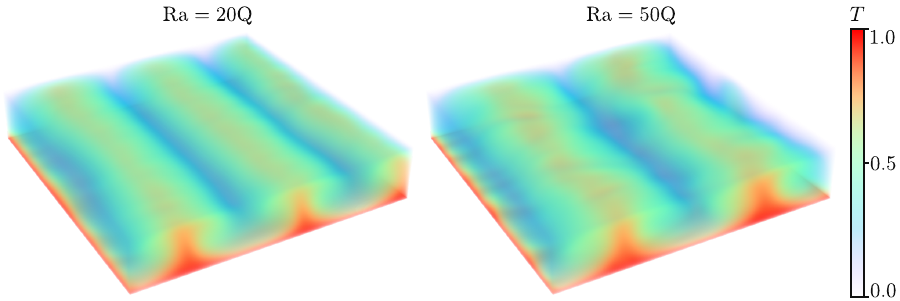}
    \caption{MHD-RBC temperature fields stabilized by the final BPTT policies.}
    \label{fig:RBCMHD_renders}
\end{figure}
\begin{wraptable}[9]{r}{0.49\textwidth}
    % \vspace{-15pt}
    \centering
    \caption{Interquartile mean reward over 10 test episodes across regimes (best method in \textbf{bold}).}
    \label{tab:test_results}
    \input{tab/test/RBCMHD3D_test_reward}
\end{wraptable}
\begin{wrapfigure}{R}{0.49\textwidth}
    \centering
    \includegraphics[width=0.49\textwidth]{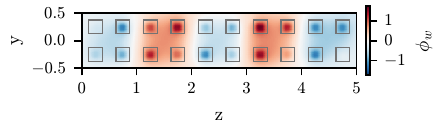}
    \caption{BPTT policy producing five convection rolls, visualized through the wall potential $\phi_w$.}
    \label{fig:rbc_mhd_actuation}
\end{wrapfigure}
This is consistent with the final evaluation of policies on the test set (see Table~\ref{tab:test_results}).
Over $10$ test episodes, BPTT achieves the highest mean reward across all methods and regimes.
Qualitative temperature fields generated by the final BPTT policy are shown in Figure~\ref{fig:RBCMHD_renders}.
Notably, for the $\rayleigh = 20\chandr$ regime, BPTT is able to stabilize five convection rolls, improving heat transfer by over $25\%$ throughout testing.
For the $\rayleigh = 50\chandr$ regime, quantitative results show more than a $15\%$ increase in $\nusselt$.
However, due to the higher Rayleigh number and associated turbulence, stabilizing the system becomes substantially harder in this regime.
These findings align with the final temperature fields controlled by BPTT, where four instead of five rolls are stabilized by the controller compared to the $\rayleigh = 20\chandr$ regime.

When inspecting the actions applied by the final BPTT policy, we observe current injection related to the five convection rolls.
Figure~\ref{fig:rbc_mhd_actuation} shows the wall potential at the Hartmann wall at $x = 0$, where actuators are visualized through gray rectangles.
Notably, we observe five large patches of alternating positive and negative electric potential spanning the wall from $z=0$ to $z=5$, likely injecting current in a way that produces a Lorentz force stabilizing the convection rolls.
These results further suggest that control policies learned through the differentiable physics systematically leverage the underlying equations to stabilize the system and improve its overall heat transfer.
Detailed qualitative and quantitative results across all algorithms are stated in Appendix~\ref{sec:app:results}.

\section{Inverse Design: Detecting Defective Wall Patches in a Convection Box}

\subsection{Experimental Setup}

\begin{wrapfigure}[17]{R}{0.49\textwidth}
    \centering
    \includegraphics[width=\linewidth]{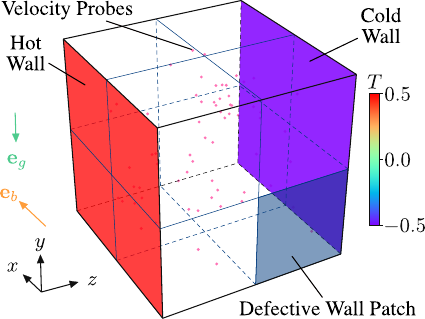}
    \caption{Defective wall patch detection setup. Based on sparse sensor probes, the goal is to recover defective wall patches in a convective box.}
    \label{fig:inverse_cw_setup}
\end{wrapfigure}

In the following, we consider an inverse design problem in a 3D buoyant liquid metal flow in a cube~\citep{tagawa2002}.
Motivated by the idea that cracks in the insulating coatings of fusion blankets can locally short-circuit the electric currents, the task aims to detect such effects from sparse velocity measurements in the bulk of the flow (see Figure~\ref{fig:inverse_cw_setup}).
The walls are divided into 24 patches, each with its own wall conductance ratio $C_w$, and the task is to recover this $C_w$ map solely from a small number of velocity probes in the bulk of the flow.
Each candidate map is evaluated by simulating the flow and comparing its probe signals with those produced by the hidden ground truth.
In addition to black-box optimization, the differentiability of \solver{} provides the gradient of this mismatch with respect to all patches, which enables gradient-based optimization.

\paragraph{Flow Configuration.}
The flow setup follows the convection benchmark of \citet{tagawa2002}, subject to a horizontal magnetic field $\eb = \bm{e}_x$ and gravity $\egrav = -\bm{e}_y$.
The dimensionless parameters are set to $\pr = 0.0321$, $\gr = 3.115 \times 10^6$, and $\ha = 100$.
Simulations are conducted in the unit cube of size $L_x \times L_y \times L_z = 1 \times 1 \times 1$.
A hot wall with constant temperature $\temp = 0.5$ is prescribed at $z=-0.5$, a cold 
wall with $\temp = -0.5$ is set at $z = 0.5$.
Temperatures are relative to a reference temperature $\temp_0 = 0$, and the remaining walls are adiabatic.
All walls are no-slip.
Using the thin-wall condition, a fixed $C_w$ is prescribed per patch.
The grid has 64 cells per dimension, i.e., $64 \times 64 \times 64$ cells in total, with $\tanh$  refinement (see \citet{krasnov2023} for details) towards all walls.
We use a fixed time step of $\Delta t = 2 \times 10^{-4}$ free-fall time units.
All runs start from the same statistically steady flow $\bm{u}_\mathrm{start}$ obtained with a uniform $C_w = 0.1$.

\paragraph{Parameterization and Target.}
Each of the six cube surfaces is equally divided into $2 \times 2$ patches, resulting in 24 design parameters $\bm{p} = (\log_{10} C_w^0, \dots, \log_{10} C_w^{23})$, bounded by $C_w \in [10^{-3}, 50]$.
A random seed defines the hidden ground-truth map:
Each face receives a base value $\log_{10} C_w \sim \mathcal{U}(-2, 0)$ shared by its four patches, and each patch is independently marked as \emph{defective} with probability $0.25$, which raises its $\log_{10} C_w$ by an offset drawn from $\mathcal{U}(0.5, 1.5)$.
Simulating this map for $N = 100$ steps from $\bm{u}_\mathrm{start}$ yields the target flow $\bm{u}_\mathrm{target}$.
Starting from random maps drawn from the same distribution, the optimizers have no prior information about defect locations.

\paragraph{Sensors and Objective.}
The flow is observed only through $64$ point probes, each measuring all three velocity components in a single cell, i.e., $192$ measurements in total.
The probe cells are drawn once, uniformly at random from the core of the domain at a distance of at least $0.1$ from all walls, and are kept fixed across all seeds and optimizers.
With the sensor operator $\mathcal{O}$, each evaluation of a candidate $\bm{p}$ restarts from an initial, steady-state velocity field $\bm{u}_\mathrm{start}$, simulates the same $N$ steps to obtain $\bm{u}_N(\bm{p})$, and computes the objective measuring the deviation from the target measurements:
\begin{equation}
    J(\bm{p}) = \frac{\lVert \mathcal{O}(\bm{u}_N(\bm{p})) - \mathcal{O}(\bm{u}_\mathrm{target}) \rVert_2^2}{\lVert \mathcal{O}(\bm{u}_\mathrm{start}) - \mathcal{O}(\bm{u}_\mathrm{target}) \rVert_2^2}.
\end{equation}

\paragraph{Optimization Algorithms.}
For this task, we consider the black-box optimization algorithm CMA-ES~\citep{hansen2001} with an initial step size of $\sigma_0 = 0.5$ and its default population size as a baseline.
Further, we evaluate the gradient-based L-BFGS-B algorithm~\citep{byrd1995} implemented in SciPy~\citep{virtanen2020}.
An optimization run stops after reaching $J \leq 10^{-5}$ or after a wall-time limit of $30$ minutes.
Detailed hyperparameters are stated in Appendix~\ref{sec:app:experiments:optim_hps}.

\subsection{Results}
\begin{wrapfigure}[15]{R}{0.49\textwidth}
    \vspace{0.78em}
    \centering
    \includegraphics[width=\linewidth]{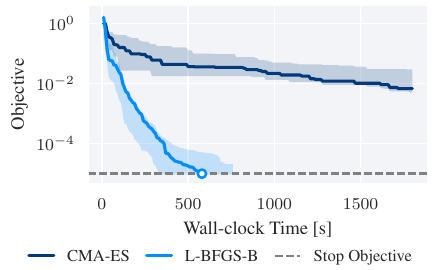}
    \caption{Defect detection optimization using CMA-ES and L-BFGS-B: best objective $J$ over wall-clock time. Lines and shading show median and min–max range across seeds, respectively.}
    \label{fig:optimization_run}
\end{wrapfigure}
In the following, we discuss the optimization results of CMA-ES and L-BFGS-B on the defective detection task.
The optimization progress over time is shown in Figure~\ref{fig:optimization_run}.
We compare the methods in terms of the objective $J$ over the wall-clock time.
Dots mark where the median reaches the stop objective $J = 10^{-5}$.
Notably, within the given time budget, the gradient-based L-BFGS-B reaches an objective value approximately $3$ orders of magnitude smaller than CMA-ES.
Similarly, the RMSE between the optimized and ground-truth wall patch conductivities is substantially smaller for the gradient-based optimizer.
These results demonstrate the efficiency of gradient-based optimization via differentiable simulation for inverse problems.
\begin{figure}[tpb]
    \centering
    \includegraphics[width=\textwidth]{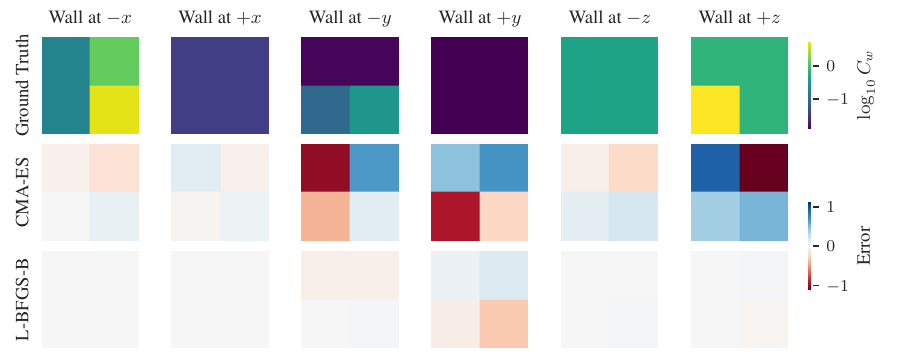}
    \caption{Final wall conductance ratio maps for CMA-ES and L-BFGS-B (seed $0$). We visualize the ground-truth $C_w$ maps per cube wall and the respective errors of the final results of both optimizers.}
    \label{fig:cw_patches_0}
\end{figure}

\noindent
When inspecting the final obtained wall patch conductivities by the optimizers, we observe notable differences.
Figure~\ref{fig:cw_patches_0} shows the ground-truth $C_w$ maps per wall as well as the respective errors of the best configuration returned by each optimizer.
For the walls at $\pm x$, both optimizer errors are comparably low.
However, for the remaining four walls, CMA-ES exhibits substantial errors, whereas the errors of L-BFGS-B remain small.
Most importantly, for the walls $\pm x$ and $\pm z$, the errors of L-BFGS-B are negligible.
Together, these results highlight that differentiating through \solver{} yields informative gradients, which enable gradient-based optimizers to recover the true underlying defective patches accurately.
Across additional seeds, residual errors vary across walls without favoring specific coordinates, which indicates this localized variation is not systematic.

\begin{figure}[tpb]
    \centering
    \includegraphics[width=\textwidth]{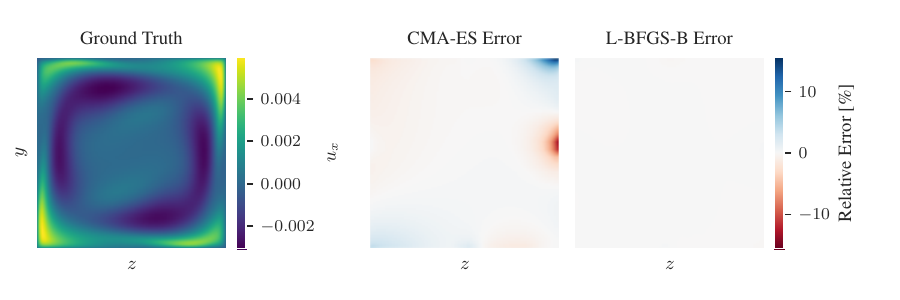}
    \caption{Ground-truth velocity fields and relative errors of velocity fields produced by the final $C_w$ maps optimized via CMA-ES and L-BFGS-B.}
    \label{fig:velocity_0}
\end{figure}

Similarly, the final velocity fields generated with the optimized $C_w$ maps differ in the errors with respect to the ground-truth velocity field.
As shown in Figure~\ref{fig:velocity_0}, the velocity field produced by the CMA-ES $C_w$ map shows relative errors on the order of $10\%$, while the L-BFGS-B velocity field errors are negligible.
This comparison demonstrates that the solution obtained through differentiable physics is robust and accurately recovers the underlying physical properties of the target flow. 

\section{Limitations \& Future Work}
In this work, we present \solver{}, a differentiable GPU-accelerated solver for liquid metal MHD flows targeted at control and inverse design problems.
Despite its GPU acceleration and efficiency optimizations, computational limitations remain.
Currently, the solver requires a CUDA-capable GPU and is restricted to single-GPU execution, though scaling to multi-GPU architectures presents a viable direction for future research.
Furthermore, high-resolution 3D simulations remain computationally demanding and highlight the need for surrogate modeling approaches.

Beyond addressing the aforementioned limitations, \solver{} can serve as an efficient data generator to train fast surrogate models, ranging from machine learning surrogates to physics foundation models~\citep{wu2024a,alkin2024,holzschuh2025} and possibly future models targeted at controlled flows.
While combining RL and differentiable physics has proven highly sample-efficient in robotics \citep{xu2021,xing2025}, its application to fluids remains largely unaddressed, a gap our benchmark aims to bridge.
Finally, we plan to extend this framework to more control and inverse problems spanning both hydrodynamics and magnetohydrodynamics.

\newpage
\subsection*{Generative AI Use}
Generative AI tools were used for implementation, code optimization, and refactoring, as well as for improving the written text of the manuscript.
The authors reviewed and verified all AI-assisted content and take responsibility for the final content of the paper.

\acks{JB and SP acknowledge funding from the European Research Council (ERC Starting Grant ``KoOpeRaDE'') under the European Union’s Horizon 2020 research and innovation programme (Grant agreement No. 101161457).
The computations were performed on the compute cluster of the Lamarr Institute for Machine Learning and Artificial Intelligence, as well as on the high-performance computers ``Noctua 2'' and ``Otus'' at the NHR Center Paderborn Center for Parallel Computing (PC2), both of which are funded by the Federal Ministry of Research, Technology and Space and by the state of North Rhine-Westphalia. The work of DK and JS is supported by the Deutsche Forschungsgemeinschaft.}

\bibliography{main}

\newpage
\appendix
\section{Differentiable Simulation of Magnetoconvective Fluid Flows}
\label{sec:app:solver}
\input{content/app_solver}

\newpage
\section{Solver Validation}
\label{sec:app:validation}
\input{content/app_validation}

\newpage
\section{Experiments}
\label{sec:app:experiments}
\input{content/app_experiments}

\newpage
\section{Results}
\label{sec:app:results}
\input{content/app_results}

\end{document}

%% file: math_commands.tex
\usepackage{amsmath,amsfonts,bm}

\def\twofigref#1#2{Figures \ref{#1} and \ref{#2}}

\def\eqref#1{equation~\ref{#1}}
\def\1{\bm{1}}

\DeclareMathAlphabet{\mathsfit}{\encodingdefault}{\sfdefault}{m}{sl}
\SetMathAlphabet{\mathsfit}{bold}{\encodingdefault}{\sfdefault}{bx}{n}

%% file: math_definitions.tex
\usepackage{amsmath,amsfonts,bm}

\newcommand{\gradfn}[2]{\partial #1 \left( \partial #2 \right)}

\newcommand{\transform}{{\mathbf{T}^{-1}}}
\newcommand{\transforminv}{\mathbf{T}}
\newcommand{\transformJ}{J}

\newcommand{\vel}{\mathrm{\mathbf{u}}}

\newcommand{\guess}{*}

\newcommand{\PMat}{P}

\newcommand{\facesign}{N}
\newcommand{\normal}{\mathbf{n}}
\newcommand{\cell}{P}
\newcommand{\nb}{F}
\newcommand{\area}{a}
\newcommand{\Snb}{\textbf{\nb}}
\newcommand{\nbf}{f}
\newcommand{\Snbf}{\textbf{\nbf}}
\newcommand{\nbb}{B}
\newcommand{\Snbb}{\textbf{\nbb}}
\newcommand{\nbfb}{b}
\newcommand{\Snbfb}{\textbf{\nbfb}}
\newcommand{\tangentialDir}{{D}}
\newcommand{\nbd}{{\tangentialDir_\nb}}
\newcommand{\Snbd}{{\textbf{\tangentialDir}_\nb}}

\newcommand{\nbdfb}{{\tangentialDir_\nbfb}}
\newcommand{\Snbdfb}{{\textbf{\tangentialDir}_\nbfb}}

\newcommand{\evalloc}[2]{{\left[ #1 \right]_{#2}}}
\newcommand{\evalcell}[1]{\evalloc{#1}{\cell}}
\newcommand{\evalnb}[1]{\evalloc{#1}{\nb}}

\newcommand{\evalnbf}[1]{\evalloc{#1}{\nbf}}

\newcommand{\lerp}[1]{\overline{#1}}

\newcommand{\ha}{\mathrm{Ha}}
\newcommand{\reyn}{\mathrm{Re}}
\newcommand{\rayleigh}{\mathrm{Ra}}
\newcommand{\stuart}{\mathrm{N}}
\newcommand{\gr}{\mathrm{Gr}}
\newcommand{\pr}{\mathrm{Pr}}
\newcommand{\chandr}{\mathrm{Q}}

\newcommand{\epot}{\phi}
\newcommand{\currDens}{\mathrm{\mathbf{j}}}

\newcommand{\currDensFlux}{\mathcal{X}}
\newcommand{\magn}{\mathbf{B}}

\newcommand{\uxB}{\vel \times \magn}
\newcommand{\eb}{\mathrm{\mathbf{e}}_b}
\newcommand{\lorentz}{\mathrm{\mathbf{F}}_L}
\newcommand{\EpotMat}{\Phi}
\newcommand{\epotRHSvec}{\mathrm{\mathbf{g}}}

\newcommand{\epotRHS}{g}

\newcommand{\gradfnCell}[2]{\partial #1 \left( \evalcell{\partial #2} \right)}
\newcommand{\gradfnNb}[2]{\partial #1 \left( \evalnb{\partial #2} \right)}

\newcommand{\egrav}{\mathrm{\mathbf{e}}_g}
\newcommand{\temp}{\mathrm{T}}
\newcommand{\buoyancy}{\mathrm{\mathbf{F}}_B}

\newcommand{\MAE}{\epsilon_\text{MAE}}
\newcommand{\MAEbf}{\boldsymbol{\epsilon}_\textbf{MAE}}
\newcommand{\RMSE}{\epsilon_\text{RMSE}}
\newcommand{\RMSEbf}{\boldsymbol{\epsilon}_\textbf{RMSE}}

\newcommand{\nusselt}{\mathrm{Nu}}

%% file: tab/test/RBCMHD3D_test_reward.tex
\begin{tabular}{lccc}
\toprule
$\rayleigh$ & SAC & BPTT & SHAC \\
\midrule
$20\chandr$ & $0.133$ & $\mathbf{0.236}$ & $0.146$ \\
$50\chandr$ & $-0.018$ & $\mathbf{0.116}$ & $0.033$ \\
\bottomrule
\end{tabular}

%% file: content/app_solver.tex
\subsection{Governing Equations}
Adapting the formulation of~\citet{masdelesvalls2011}, we extend the PISO algorithm implemented in PICT~\citep{franz2026} to $\epot$-PISO.
From the charge conservation in Equation~\ref{eq:current_density_charge_conservaton_lorentz_force}, we can derive the Poisson equation for the electric potential $\epot$:
\begin{align}\label{eq:app:poisson_phi}
    \nabla \cdot \currDens = 0 \nonumber \\
    \Leftrightarrow \nabla \cdot (- \nabla \epot + \vel \times \eb) = 0 \nonumber \\
    \Leftrightarrow - \nabla \cdot \nabla \epot + \nabla \cdot (\vel \times \eb) = 0 \nonumber \\
    \Leftrightarrow \nabla^2 \epot = \nabla \cdot (\vel \times \eb)  \, .
\end{align}

\paragraph{Boundary Conditions}
For the electric potential $\epot$, boundary conditions are defined by the wall conductance ratio 
\begin{align}
    C_w = \frac{\sigma_w \tau_w}{\sigma H}
\end{align}
with
\begin{align}
    C_w &= 0 \implies \left. \frac{\partial \epot}{\partial n} \right|_w = 0 \quad &\text{(Perfectly insulating wall)} \\
    C_w &\rightarrow \infty \implies \epot = \text{const.} \quad &\text{(Perfectly conducting wall)}
\end{align}

As formulated by \citet{walker1981}, for walls with a constant conduction ratio $C_w >0$ and wall thickness $\tau \ll \mathrm{H}$, we employ the so called \emph{thin-wall boundary condition}
\begin{align}
    \left. \frac{\partial \epot}{\partial n} \right|_w = C_w \left.\nabla^2_\tau \epot \right|_w \,.
\end{align}

% To ensure numerical stability and charge conservation, we employ the scheme of~\citet{ni2007}~\citep[cf.][]{masdelesvalls2011,tassone2016}, which enforces a conservative discretization of the Lorentz force.

\subsection{Finite Volume Formulation}

\paragraph{Notation}
For consistency and direct compatibility with the underlying solver, we adopt the finite volume method~\citep[cf.][]{maliska2023} notation used by~\citet{franz2026}, summarized below:

{
\centering
\begin{tabular}{rl}
    Current cell & $\cell$ \\
    Set of valid neighbor cells of $\cell$ & $\nb \in \Snb$ \\
    Set of faces between $\cell$ and $\Snb$ & $\nbf \in \Snbf$ \\
    Set of (virtual) boundary neighbor cells of $\cell$ & $\nbb \in \Snbb$ \\
    Set of boundary-faces between $\cell$ and $\Snbb$ & $\nbfb \in \Snbfb$ \\
    Set of diagonal neighbors 'tangential' to $\nb$ & $\nbd \in \Snbd$\\
    Set of diagonal boundary faces 'tangential' to $\nbfb$ & $\nbdfb \in \Snbdfb$\\
    Sign of logical face direction on computational grid & $\facesign_\nbf$ \\
    Evaluated/value at current cell & $\evalcell{\Box}$ or $\Box_\cell$ \\
    Evaluated/value at neighbor cell & $\evalnb{\Box}$ or $\Box_\nb$ \\
    (Linear) Interpolation & $\lerp{\Box}$
\end{tabular}
}

\newpage
Following this convention, for cell $\cell$ (index $i$) and neighbor cell $\nb \in \Snb$ (index $j$), the entries of a discretized operator matrix $\EpotMat$ are indexed as:

{
\centering
\begin{tabular}{rl}
    Diagonal entry at current cell & $\evalcell{\EpotMat_\cell} = \EpotMat_{ii}$ \\
    Off-diagonal entry for neighbor & $\evalcell{\EpotMat_\nb} = \EpotMat_{ij}$ \\
    Neighbor row, current column & $\evalnb{\EpotMat_\cell} = \EpotMat_{ji}$ \\
    Neighbor row diagonal entry & $\evalnb{\EpotMat_\nb} = \EpotMat_{jj}$
\end{tabular}
\par
}

\paragraph{Potential Solve}

To solve the Poisson equation for the electric potential ($\nabla^2\epot^\guess = \nabla \cdot (\uxB)$), we leverage the Conjugate Gradient (CG) solver in PICT.
The discrete potential matrix $\EpotMat$ is constructed similarly to the discrete pressure matrix $\PMat$, but the diagonal entries $A^{-1}$ are omitted.
This leaves us with the Laplace coefficients $\evalnbf{ \lerp{ \alpha_{jj} }}$:

\begin{equation} \label{eq:app:epotMatDiscrete}
    \begin{split}
        \evalcell{ \EpotMat_\nb } &= \evalnbf{ \lerp{ \alpha_{jj} }} \\
        \evalcell{ \EpotMat_\cell } &= \sum_{\Snbf} - \evalnbf{ \lerp{ \alpha_{jj} }} \,.
    \end{split}
\end{equation}
Non-orthogonal grid corrections follow the same formulation as the pressure equation.

Applying the divergence theorem, the right-hand side of the potential equation is discretized as
\begin{align*}
    \evalcell{ \nabla \cdot (\uxB) } \approx \sum_{\Snbf} (\uxB)_{\nbf} \cdot \normal_{\nbf} \area_{\nbf} \,. 
\end{align*}

In the generalized coordinate systems, this can be expressed as
\begin{equation} \label{eq:app:epotRHSdivDiscrete}
    \evalcell{ \nabla \cdot (\uxB) } = \sum_{\Snb \rightarrow j} \evalnbf{ \lerp{ \transformJ \transforminv_j \cdot (\uxB) } } \facesign_\nb \, .
\end{equation}

\paragraph{Face-based Lorentz Force}

To ensure discrete charge conservation, we compute the Lorentz force from face-centered current density fluxes $\currDensFlux$ following the conservative scheme of~\citet{ni2007}:
\begin{equation} \label{eq:app:currDensFlux}
    \currDensFlux_\nbf^j = - \underbrace{\evalnbf{ \lerp{ \alpha_{jj} }} \facesign_\nb (\evalnb{\epot} - \evalcell{ \epot })}_{\evalnbf{ \nabla \epot }} + \evalnbf{ \lerp{ \transformJ \transforminv_j \cdot (\uxB) } } \,.
\end{equation}

The cell-centered current density $\currDens_\cell$ is obtained by averaging opposite face fluxes along each coordinate direction $j$ and transforming to the physical domain:
\begin{equation} \label{eq:app:currDens}
    \currDens_\cell = \frac{1}{\transformJ} \sum_j \left(\transform \right)_{\cdot j} \left(\frac{1}{2} \sum_{\nb \rightarrow j} \currDensFlux_\nbf^j \right) \, .
\end{equation}
The resulting Lorentz body force is then evaluated as $\lorentz = \stuart (\currDens_\cell \times \magn)$.

\paragraph{Thin-Wall Approximation}
For walls with arbitrary conductance ratio $C_w$, we apply the thin-wall boundary condition.
For a fluid cell $\cell$ adjacent to a boundary face $\nbfb \in \Snbfb$, this condition couples the potential to the tangential neighbor cells $\nbd \in \Snbd$.
For each tangential direction $\tau \neq n$, we define the surface Laplace coefficient as:
\begin{equation}\label{eq:sigma_tau_nbf}
\sigma_{\tau, \nbf} = C_w \cdot \frac{1}{2} \left( \evalcell{ \alpha_{\tau\tau} } |\mathbf{T}^{-1}_{n, \cell}| + \evalnb{ \alpha_{\tau\tau} } |\mathbf{T}^{-1}_{n, \nb}| \right) \,,
\end{equation}
where $\alpha_{\tau\tau}$ denotes the orthogonal Laplace coefficient along $\tau$.

The discrete potential matrix $\EpotMat$ in Equation~\ref{eq:app:epotMatDiscrete} is updated accordingly:
\begin{equation}
    \begin{split}
    \evalcell{ \EpotMat_\cell } & \leftarrow \evalcell{ \EpotMat_\cell } - \sigma_{\tau, \nbf} \\
    \evalcell{ \EpotMat_{\nbd} } & \leftarrow \evalcell{ \EpotMat_{\nbd} } + \sigma_{\tau, \nbf} \,.
    \end{split}
\end{equation}
while the wall-normal face contribution remains zero as in the insulating case.
For cells bordered by multiple thin-wall boundaries (such as corner cells), these corrections accumulate additively across all adjacent boundary faces $\nbfb \in \Snbfb$.

\subsection{Gradients}
\label{sec:app:solver:gradients}
For gradient computation, we adopt the notation of \citet{franz2026}.
In this work, $\partial y := \left(\frac{\partial L}{\partial y}\right)^{\!\top}$ denotes the adjoint, i.e., the gradient of a scalar loss $L$ with respect to $y$, rather than a differential.
For a function $y = h(x)$, the vector-Jacobian product $\gradfn{x}{y}_f = \left(\frac{\partial f}{\partial x}\right)^{\!\top} \partial y$ yields the contribution to $\partial x$ that flows through $y$ and has the same shape as $x$.
If $x$ influences the loss through multiple quantities, the individual contributions are summed to obtain $\partial x$.
We denote the function by $h$ to avoid confusion with the face index $\nbf$.
Standard operations handled automatically by PyTorch's \texttt{autograd} are omitted.

\paragraph{Local Formulation.}
$x$ and $y$ denote arbitrary physical quantities in the computational graph, e.g., the potential $\epot$ or the velocity $\vel$.
As in \citet{franz2026}, we write them per cell or face: $\evalcell{\epot}$ is considered a scalar and $\evalcell{\vel}$ a vector of three components, although the corresponding fields are vectors over all cells. 
Since every operation, except for the linear solves, only couples a cell to its neighbors, the global Jacobian $\partial h / \partial x$ is sparse and never assembled. 
Instead, we write the vector-Jacobian product for a single cell $\cell$.
Since $\evalcell{x}$ only enters quantities of $\cell$ itself and of its neighbors $\nb \in \Snb$, all other terms of the global product vanish.
For a cell-based quantity $y$, we have
\begin{align}
    \gradfn{\evalcell{x}}{y}
    = \left(\frac{\partial \evalcell{y}}{\partial \evalcell{x}}\right)^{\!\top} \evalcell{\partial y}
    + \sum_{\nb \in \Snb} \left(\frac{\partial \evalnb{y}}{\partial \evalcell{x}}\right)^{\!\top} \evalnb{\partial y} \, ,
\end{align}
and analogously, for a face-based quantity $y$, a sum over the bounding faces $\nbf \in \Snbf$:
\begin{align}
    \gradfn{\evalcell{x}}{y}
    = \sum_{\nbf \in \Snbf} \left(\frac{\partial \evalnbf{y}}{\partial \evalcell{x}}\right)^{\!\top} \evalnbf{\partial y} \, .
\end{align}
Inserting the local forward rule for $y$ turns this into an explicit per-cell expression that maps directly onto a GPU kernel.

\paragraph{Overview of the Backward Pass.}
The velocity $\vel$ affects the Lorentz force through $\uxB$ along two paths:
directly via the face current flux, and indirectly via the electric potential $\epot$,
whose Poisson equation has the right-hand side $\epotRHS = \nabla\cdot(\uxB)$.
Starting from the downstream gradient $\partial\lorentz$, the backward pass proceeds as follows:
\begin{enumerate}
    \item Lorentz force:
          $\partial\currDens = \frac{\ha^2}{\reyn}\, \magn \times \partial\lorentz$.
    \item Cell current density to face fluxes:
          $\partial\currDens \rightarrow \partial\currDensFlux_\nbf^j$
          (Eq.~\ref{eq:app:currDensFluxGrad}).
    \item Face fluxes to potential:
          $\partial\currDensFlux_\nbf^j \rightarrow \partial\epot$
          (Eq.~\ref{eq:app:epotGrad}).
    \item Face fluxes to $\uxB$, \emph{direct path}:
          $\partial\currDensFlux_\nbf^j \rightarrow \partial(\uxB)$
          (Eq.~\ref{eq:app:uxBGrad}).
    \item Potential to $\uxB$, \emph{indirect path}: adjoint solve
          $\EpotMat^T \partial\epotRHS = \partial\epot$, followed by the backward
          divergence $\partial\epotRHS \rightarrow \partial(\uxB)$
          (Paragraph Potential Solve).
    \item Both contributions to $\partial(\uxB)$ from steps 4 and 5 are summed.
    \item Cross product:
          $\partial\vel = \magn \times \partial(\uxB)$.
\end{enumerate}
The following paragraphs derive steps 2 -- 5 in detail.

\paragraph{Face-based Lorentz Force.}
We trace the analytical backward pass starting from the Lorentz body force $\lorentz = \frac{\ha^2}{\reyn}(\currDens \times \magn)$.
Backpropagating through the face-averaged current density in Equation~\ref{eq:app:currDens} yields the face flux sensitivity:
\begin{align} \label{eq:app:currDensFluxGrad}
    \gradfn{\currDensFlux_\nbf^j}{\currDens} = \sum_i \left( \frac{1}{2 \transformJ} \left(\transform \right)_{ij} \partial \currDens^i \right) \, .
\end{align}
The cell gradient $\evalcell{\partial \currDens}$ is thus distributed across all bounding faces $\nbf \in \Snbf$ via $\partial \currDensFlux_\nbf^j$.

Substituting the potential difference $x_\nbf^j = \evalnb{\epot} - \evalcell{\epot}$, the face current flux formulation (see Equation~\ref{eq:app:currDensFlux}) simplifies to
\begin{align}
    \currDensFlux_\nbf^j = - \evalnbf{ \lerp{ \alpha_{jj} }} \facesign_\nbf x_\nbf^j + \evalnbf{ \lerp{ \transformJ \transforminv_j \cdot (\uxB) } } \, .
\end{align}
with sensitivity with respect to $x_\nbf^j$:
\begin{align}
    \gradfn{x_\nbf^j}{\currDensFlux_\nbf^j} = - \evalnbf{ \lerp{ \alpha_{jj} }} \facesign_\nb \partial \currDensFlux_\nbf^j \, .
\end{align}
Because the flux at face $\nbf$ couples both cell $\cell$ and neighbor $\nb$, the individual derivatives are:
\begin{align}
    \begin{split}
        \gradfn{\evalcell{\epot}}{\currDensFlux_\nbf^j} &= - \gradfn{x_j}{\currDensFlux_\nbf^j} = \evalnbf{ \lerp{ \alpha_{jj} }} \facesign_\nbf \partial \currDensFlux_\nbf^j \\
        \gradfn{\evalnb{\epot}}{\currDensFlux_\nbf^j} &= + \gradfn{x_j}{\currDensFlux_\nbf^j} = -\evalnbf{ \lerp{ \alpha_{jj} }} \facesign_\nbf \partial \currDensFlux_\nbf^j \, .
    \end{split}
\end{align}
Consequently, each face $\nbf$ receives two gradient contributions:
\begin{enumerate}
    \item from the cell's current density gradient $\evalcell{\partial \currDens}$, denoted as $\evalcell{\partial \currDensFlux_\nbf^j} := \gradfnCell{\currDensFlux_\nbf^j}{\currDens}$, and
    \item from neighbor cell $\nb$'s current density gradient $\evalnb{\partial \currDens}$, denoted as $\evalnb{\partial \currDensFlux_\nbf^j} := \gradfnNb{\currDensFlux_\nbf^j}{\currDens}$.
\end{enumerate}
Accounting for the reversed face normal sign ($\facesign'_\nbf = -\facesign_\nbf$) from neighbor $\nb$, the total potential gradient for cell $\cell$ is:
\begin{align} \label{eq:app:epotGrad}
    \gradfn{\evalcell{\epot}}{\currDensFlux_\nbf^j} &= \sum_{\Snb \rightarrow j} \left[ \gradfnCell{\evalcell{\epot}}{\currDensFlux_\nbf^j} + \gradfnNb{\evalcell{\epot}}{\currDensFlux_\nbf^j} \right] \nonumber \\
    &= \sum_{\Snb \rightarrow j} \left[ \evalnbf{ \lerp{ \alpha_{jj} }} \facesign_\nbf \evalcell{\partial \currDensFlux_\nbf^j} + \left( - \evalnbf{ \lerp{ \alpha_{jj} }} \facesign'_\nbf \evalnb{\partial \currDensFlux_\nbf^j} \right) \right] \nonumber \\
    &= \sum_{\Snb \rightarrow j} \evalnbf{ \lerp{ \alpha_{jj} }} \facesign_\nbf \left( \evalcell{\partial \currDensFlux_\nbf^j} + \evalnb{\partial \currDensFlux_\nbf^j} \right) \, .
\end{align}
Following the definition of $\evalnbf{\lerp{\Box}}$, we get
\begin{align*}
    \evalnbf{ \lerp{ \transformJ \transforminv_j \cdot (\uxB) } } &= \frac{\evalcell{\transformJ \transforminv_j \cdot (\uxB)} + \evalnb{\transformJ \transforminv_j \cdot (\uxB)}}{2} \,.
\end{align*}
This leads us to the gradient function
\begin{align*}
    \gradfn{\evalcell{(\uxB)}}{\currDensFlux_\nbf^j} = \frac{1}{2} \evalcell{\transformJ \transforminv_j} \partial \currDensFlux_\nbf^j \,.
\end{align*}
Accumulating contributions from adjacent cells gives the total gradient:
\begin{align} \label{eq:app:uxBGrad}
    \gradfn{\evalcell{(\uxB)}}{\currDensFlux_\nbf^j} &= \sum_{\Snb \rightarrow j} \frac{1}{2} \evalcell{\transformJ \transforminv_j} \left( \evalcell{\partial \currDensFlux_\nbf^j} + \evalnb{\partial \currDensFlux_\nbf^j} \right) \, .
\end{align}
The complete gradients $\gradfn{\evalcell{\epot}}{\currDens}$ and $\gradfn{\evalcell{(\uxB)}}{\currDens}$ follow directly by substituting the face sensitivities (see Equation~\ref{eq:app:currDensFluxGrad}) into Equation~\ref{eq:app:epotGrad} and Equation~\ref{eq:app:uxBGrad}, respectively.

\paragraph{Potential Solve}
Because the discrete potential matrix $\EpotMat$ depends purely on static grid geometry, it requires no gradient backpropagation.
Following~\citet{franz2026}, given the downstream potential gradient $\partial \epot$, we compute $\gradfn{\epotRHS}{\epot}$ by solving the linear adjoint system for the potential Poisson equation (see Equations~\ref{eq:app:poisson_phi}~and~\ref{eq:app:epotMatDiscrete}):
\begin{equation}
    \EpotMat^T \partial \epotRHS = \partial \epot \,,
\end{equation}
where $\epotRHS = \nabla \cdot (\uxB)$.
Differentiating through the discretized divergence operator in Equation~\ref{eq:app:epotRHSdivDiscrete} then yields the gradient with respect to $\uxB$:
\begin{equation}
    \gradfn{(\uxB)_\cell}{\epotRHS} = \sum_{\Snb \rightarrow j} \evalnbf{\lerp{\transformJ \transforminv_j \partial \epotRHS}} \facesign_\nb \,.
\end{equation}
The remaining gradient pathways back to the velocity field $\vel$ are handled by PyTorch's \texttt{autograd}.

\subsection{Further Improvements}
\label{sec:app:solver:further_improvements}

To assess the computational efficiency of \solver{} and quantify individual solver improvements, we benchmark throughput across the non-MHD environments in FluidGym.
For each environment, we measure throughput in environment steps per second across different configurations: unbatched PICT, \solver{} without batching, leave-one-out feature ablations, and full \solver{} across a sweep of batch sizes. 
The optimal batch size for \solver{} is automatically determined as the throughput-maximizing size that satisfies a peak GPU memory threshold (i.e., staying within a predefined percentage of device memory). 
Ablation variants are subsequently evaluated at this optimal batch size. 
The resulting speedups over original PICT, evaluated across multiple environments, are summarized in Table~\ref{tab:app:solver:solver_benchmark}.

\begin{table}[h]
    \centering
    \caption{Individual speedups per environment of \solver{} over PICT in FluidGym with optimal batch size $B$, determined on a NVIDIA A100 with 80GB VRAM. Optimal batch sizes might differ on different GPUs.}
    \label{tab:app:solver:solver_benchmark}
    \input{tab/solver_benchmark}
\end{table}

% \begin{figure}[H]
%     \centering
%     \includegraphics[width=0.49\textwidth]{gfx/plots/benchmark/solver_ablation_timing.pdf}
%     \caption{Ablation study of CG solver runtime in \solver{} and PICT. Factors show slowdowns of solvers compared to \solver{}.$^\dagger$Not all CG solves did converge in $10\,$k iterations.}
%     \label{fig:app:solver:solver_ablation_timing}
% \end{figure}

\paragraph{Linear Upwind Advection Scheme}
For the MHD flows studied in this work, we found the linear upwind differencing scheme for the advection term to be more stable than the previously implemented central difference scheme~\citep[see e.g.][]{maliska2023}.

\paragraph{Potential Solve Preconditioning}
To accelerate CG convergence, we precondition the system using Algebraic Multigrid~\citep[AMG;][]{ruge1987} via \texttt{PyAMG}~\citep{bell2023} for the electric potential.
Since the discrete potential matrix $\EpotMat$ depends only on the grid, the preconditioning can efficiently be computed once and re-used throughout the simulation.
Furthermore, AMG preconditioning is applied to decrease the number of CG iterations of the adjoint solve in the backward pass.

\paragraph{Intermediate Pressure Tolerance}
In line with \texttt{OpenFOAM}~\citep{weller1998a}, we integrate a lower intermediate tolerance for the first pressure correction step in combination with the previous final tolerance value\footnote{See \url{https://github.com/OpenFOAM/OpenFOAM-6/blob/47517f2ebb1bdbacb8e3ce4b15110cd252953f87/tutorials/combustion/reactingFoam/RAS/SandiaD_LTS/system/fvSolution\#L33} for example.}.

\paragraph{Activation Checkpointing}
To enable backpropagation of reward gradients through long horizons of, for instance, $25$ or RL steps, we implement activation checkpointing.
This allows users to trade off computational runtime and memory requirements.
In our experiments, forward and backward passes through segments of $10$ PISO steps can be computed on a GPU with $80$ GB VRAM.

%% file: tab/solver_benchmark.tex
\begin{tabular}{lcc} 
    \toprule
    Environment & $\phi$-PICT & $B$ \\
    \midrule
    Airfoil2D-easy & $\times114.07$ & 64 \\
    Airfoil2D-hard & $\times149.56$ & 96 \\
    Airfoil2D-medium & $\times114.32$ & 96 \\
    Airfoil3D-easy & $\times2.79$ & 2 \\
    Airfoil3D-hard & $\times1.77$ & 2 \\
    Airfoil3D-medium & $\times2.97$ & 4 \\
    CylinderJet2D-easy & $\times117.32$ & 384 \\
    CylinderJet2D-hard & $\times61.72$ & 128 \\
    CylinderJet2D-medium & $\times69.93$ & 192 \\
    CylinderJet3D-easy & $\times14.01$ & 24 \\
    CylinderJet3D-hard & $\times2.68$ & 2 \\
    CylinderJet3D-medium & $\times5.57$ & 12 \\
    CylinderRot2D-easy & $\times129.49$ & 384 \\
    CylinderRot2D-hard & $\times131.05$ & 512 \\
    CylinderRot2D-medium & $\times85.75$ & 192 \\
    RBC2D-easy & $\times72.95$ & 64 \\
    RBC2D-hard & $\times119.09$ & 128 \\
    RBC2D-medium & $\times116.26$ & 192 \\
    RBC2D-wide-easy & $\times62.17$ & 96 \\
    RBC2D-wide-hard & $\times80.94$ & 96 \\
    RBC2D-wide-medium & $\times80.78$ & 96 \\
    RBC3D-easy & $\times16.13$ & 96 \\
    RBC3D-hard & $\times16.81$ & 96 \\
    RBC3D-medium & $\times16.06$ & 96 \\
    RBC3D-wide-easy & $\times6.79$ & 24 \\
    RBC3D-wide-hard & $\times7.81$ & 24 \\
    RBC3D-wide-medium & $\times7.62$ & 24 \\
    TCFLarge3D-both-easy & $\times4.06$ & 48 \\
    TCFLarge3D-both-hard & $\times6.03$ & 96 \\
    TCFLarge3D-both-medium & $\times6.16$ & 96 \\
    TCFLarge3D-bottom-easy & $\times3.12$ & 32 \\
    TCFLarge3D-bottom-hard & $\times4.25$ & 96 \\
    TCFLarge3D-bottom-medium & $\times4.18$ & 96 \\
    TCFSmall3D-both-easy & $\times6.36$ & 96 \\
    \midrule
    Mean & $\times20$ &  \\
    \bottomrule
\end{tabular}

%% file: content/app_validation.tex
We validate our differentiable MHD implementation across canonical flow configurations with known analytical solutions, following standard validation protocols from prior MHD solver literature~\citep{masdelesvalls2011,tassone2016,krasnov2023,eardley-brunt2024}.
These benchmarks evaluate the solver's accuracy under varying Hartmann numbers, wall conductivity ratios, and thermal boundary conditions.

\subsection{Forward Validation}

In the following, we discuss the validation of \solver{} against canonical MHD cases.
We denote spatial coordinates as $x$, $y$, and $z$ and spatial averaging over a direction $x$ is written as $\langle \cdot \rangle_x$.

% \SP{Maybe you should briefly introduce the meaning of $x$ and $y$ here. It es equally common to use $x_1$ and $x_2$ for the spatial coordinates (but I'm fine with $x$ and $y$).}

% \SP{Have you defined the averaging notation as in, e.g., $\langle u_x\rangle_{x,y}$?}

\subsubsection{Hartmann Flow}
The Hartmann flow is a 2D flow between two infinite walls separated by a distance $2H$ exposed to a magnetic field of strength $B_0$ perpendicular to the walls.
To maintain a constant mean streamwise velocity, we dynamically update the pressure gradient $G = -\frac{dp}{dx}$ depending on the difference between the current and target velocity, i.e.,
\begin{align*}
    G^{n} \gets G^{n-1} + c (U - \langle u_x\rangle_{x,y})
\end{align*}
with $c=0.1$.
The analytical solution for the velocity profile is given as
\begin{align*}
    u_x = U \frac{\cosh (\ha) - \cosh (\ha y/H)}{\cosh (\ha) - \sinh(\ha) / \ha}
\end{align*}
with reference velocity $U$.
Following \citet{masdelesvalls2011}, we estimate the solver accuracy using the mean absolute error (MAE).
We additionally normalize by the peak analytical velocity
\begin{align*}
    \MAE = \frac{\langle |u_x - u_{x,\text{analytical}}| \rangle_{y,z}}{\max(u_{x,\text{analytical}})} \cdot 100 ,
\end{align*}
For our experiments, we set $\reyn = 10$ and $H = 1$.

\begin{figure}[H]
    \centering
    \begin{minipage}[t]{0.49\textwidth}
        \vspace{0pt}
        \centering
        \includegraphics[width=\linewidth]{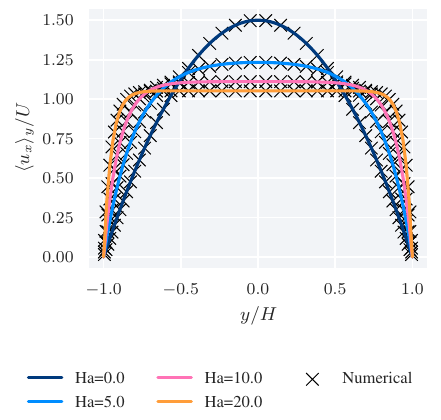}
        \caption{Hartmann case: Analytical and numerical velocity profiles for different Hartmann numbers. Numerical results are averaged over the streamwise ($x$) direction.}
        \label{fig:app:val:hartmann}
    \end{minipage}\hfill
    \begin{minipage}[t]{0.49\textwidth}
        \vspace{0pt}
        \centering
        \captionof{table}{Experimental results for the Hartmann case.}
        \label{tab:app:hartmann_error}
        \vspace{4pt}
        \input{tab/val_hartmann}
    \end{minipage}
\end{figure}

\subsubsection{Shercliff and Hunt Flows}
\begin{table}[H]
    \centering
    \caption{Experimental setup for Shercliff and Hunt cases.}
    \label{tab:app:shercliff_hunt_setup}
    \begin{tabular}{crr} 
        \toprule
        \textbf{Case} &
        $\mathrm{\mathbf{Ha}}$ &
        $\mathrm{\mathbf{Re}}$ \\
        \midrule
        \RomanNum{1} & 20 & 9 \\
        \RomanNum{2} & 100 & 10 \\
        \RomanNum{3} & 1000 & 100 \\
        \bottomrule
    \end{tabular}
\end{table}

\begin{table}[H]
    \centering
    \caption{Validation results for the Shercliff case.}
    \label{tab:app:val_shercliff}
    \input{tab/val_shercliff}
\end{table}

\begin{table}[H]
    \centering
    \caption{Validation results for the Hunt case.}
    \label{tab:app:val_hunt}
    \input{tab/val_hunt}
\end{table}

For the Shercliff and Hunt cases, the analytical profile is given by a Fourier series
\begin{align*}
    u_{x,\text{analytical}} &= \frac{U_x}{\nu}GH^2 \\
    U_x &= \sum_{k=0}^\infty \frac{2(-1)^k \cos(\alpha_k \xi)}{l \alpha^3_k} (1 - U_{2,k} - U_{3,k}) \\
    U_{2,k} &= \frac{\left( C_w r_{2,k} + \frac{1 - \exp(-2r_{2,k})}{1 + \exp(-2r_{2,k})}\right) \frac{\exp(-r_{1,k}(1-\eta)) + \exp(-r_{1,k}(1+\eta))}{2}}{\frac{1 + \exp(-2r_{1,k})}{2}C_w N_k + \frac{1 + \exp(-2(r_{1,k} + r_{2,k}))}{1 + \exp(-2r_{2,k})}} \\
    U_{3,k} &= \frac{\left( C_w r_{1,k} + \frac{1 - \exp(-2r_{1,k})}{1 + \exp(-2r_{1,k})}\right) \frac{\exp(-r_{2,k}(1-\eta)) + \exp(-r_{2,k}(1+\eta))}{2}}{\frac{1 + \exp(-2r_{2,k})}{2}C_w N_k + \frac{1 + \exp(-2(r_{1,k} + r_{2,k}))}{1 + \exp(-2r_{1,k})}} \\
    N_k &= \sqrt{\ha^2 + 4\alpha_k^2} \\
    r_{1,k}, r_{2,k} &= \frac{1}{2} \left( \pm \ha + N_k \right) \\
    \alpha_k &= \left( k + \frac{1}{2} \right) \frac{\pi}{l} \,, \\ 
\end{align*}
with $\eta = z / D$, $\xi = y / H$, and $l=H/D$.
For the Shercliff case, we set $L=2$, $H = D = 1$, and $C_w = 0$.
For the Hunt case, we set $C_w=0.1$ for the walls at $y =\pm H$.

For the Shercliff and Hunt cases, we additionally consider the cell-volume-weighted relative RMSE, denoted as
\begin{align*}
    \RMSE = \frac{\displaystyle\sqrt{\frac{\sum_i V_i \left(u_{x,i} - u_{x,\text{analytical},i}\right)^2}{\sum_i V_i}}}{\displaystyle\sqrt{\frac{\sum_i V_i\, u_{x,\text{analytical},i}^2}{\sum_i V_i}}} \cdot 100 \,,
\end{align*}
where $V_i$ denotes the volume of cell $i$ and the sums run over all cells in the domain.

\begin{figure}[H]
    \centering
    \includegraphics[width=\textwidth]{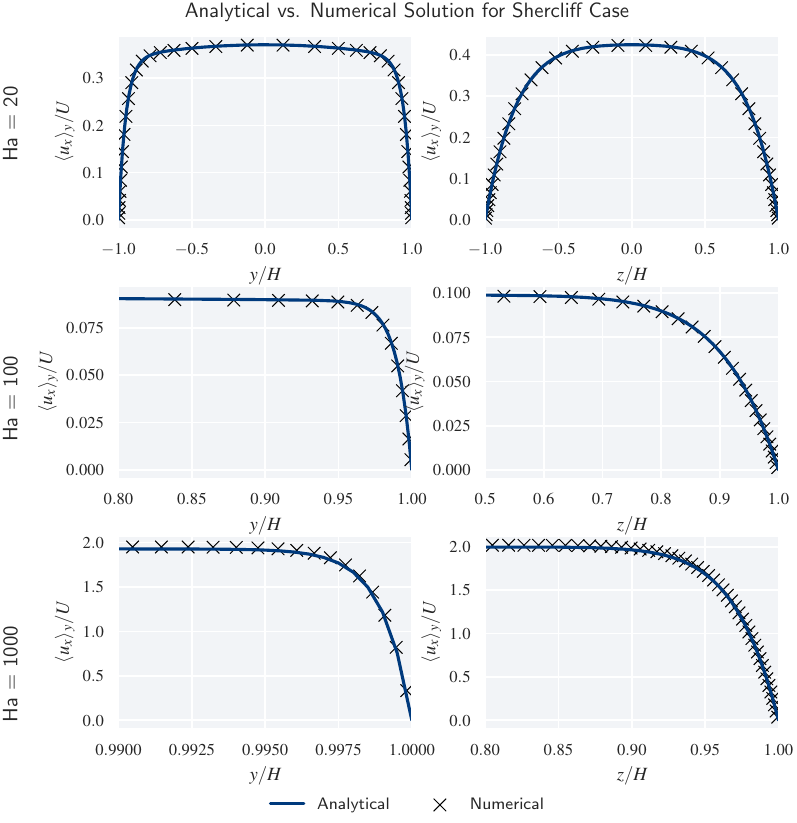}
    \caption{Shercliff case: Analytical and numerical velocity profiles for different Hartmann numbers. Numerical results are averaged over the streamwise ($x$) direction.}
    \label{fig:app:val:shercliff_profile}
\end{figure}

\begin{figure}[H]
    \centering
    \includegraphics[width=\textwidth]{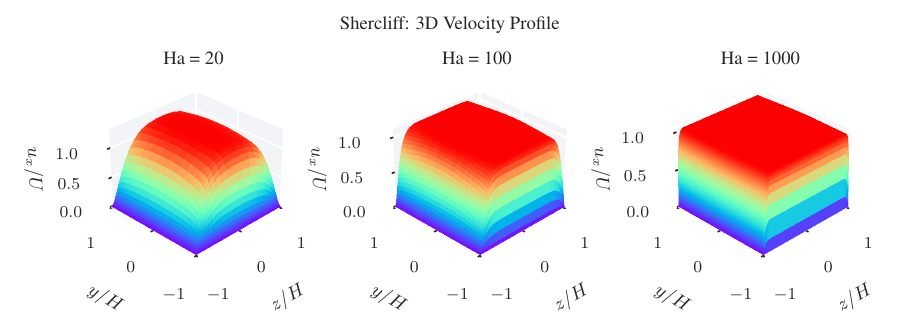}
    \caption{Shercliff case: Velocity for different Hartmann numbers.}
    \label{fig:app:val:shercliff_velocity_3d}
\end{figure}

\begin{figure}[H]
    \centering
    \includegraphics[width=\textwidth]{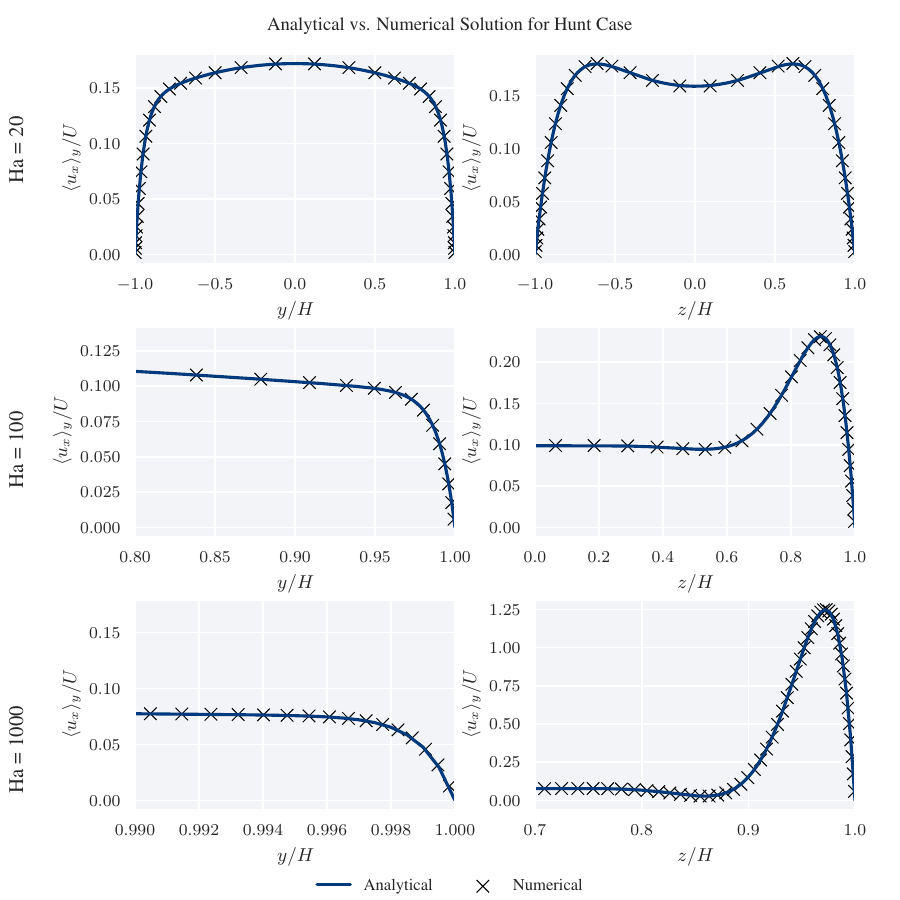}
    \caption{Hunt case: Analytical and numerical velocity profiles for different Hartmann numbers. Numerical results are averaged over the streamwise ($x$) direction.}
    \label{fig:app:val:hunt_profile}
\end{figure}

\begin{figure}[H]
    \centering
    \includegraphics[width=\textwidth]{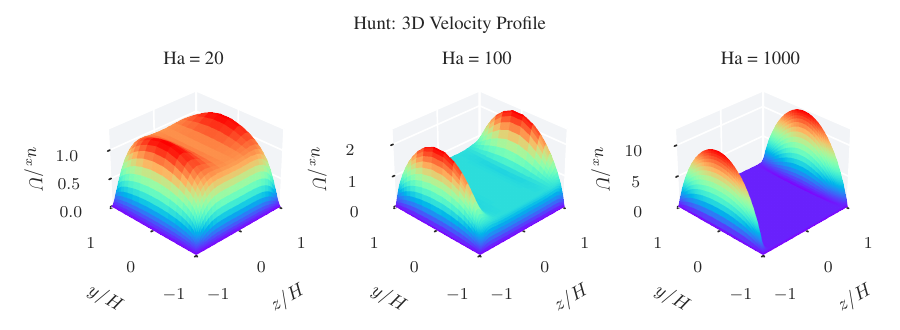}
    \caption{Hunt case: Velocity for different Hartmann numbers.}
    \label{fig:app:val:hunt_velocity_3d}
\end{figure}

\subsubsection{Jet Detachments in a Duct Flow with $\reyn = 2000$ and $\ha = 2000$.}
\begin{figure}[H]
    \centering
    \includegraphics[width=\textwidth]{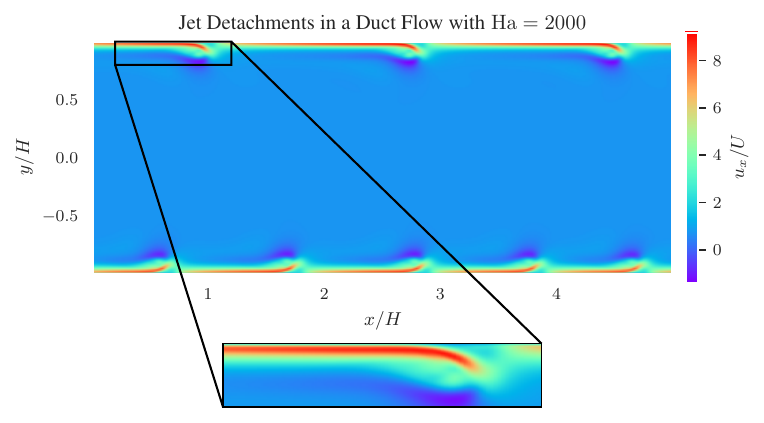}
    \caption{Jet detachments in a duct flow with $\reyn = 2000$ and $\ha = 2000$.}
    \label{fig:app:val:duct_velocity}
\end{figure}

To validate the solver in a high-$\ha$ magnetohydrodynamic regime, we simulate Hunt flow in a rectangular duct following \citet{krasnov2023} and \citet{arlt2018}.
The flow domain has dimensions $L_x \times L_y \times L_z = 5.0 \times 2.0 \times 2.0$ discretized on a grid of $N_x \times N_y \times N_z = 300 \times 140 \times 200$. 
The grid is refined towards the walls along $y$ and $z$.
Periodic boundary conditions are applied along the streamwise $x$-direction.
The flow is governed by $\mathrm{Re} = 2000$ and $\mathrm{Ha} = 2000$, with an external magnetic field oriented along the $z$-axis.
The duct walls parallel to the magnetic field are electrically insulated ($C_{w,\parallel} = 0$), whereas the perpendicular (Hartmann) walls have a finite conductance ratio ($C_{w,\perp} = 0.03$).
Under these conditions, the fully developed regime exhibits characteristic high-velocity side-layer jets and jet detachment phenomena as described by \citet{arlt2018} and \citet{krasnov2023} (see \twofigref{fig:app:val:duct_velocity}{fig:app:val:duct_u_x_over_x})

\begin{figure}[H]
    \centering
    \includegraphics[width=0.49\textwidth]{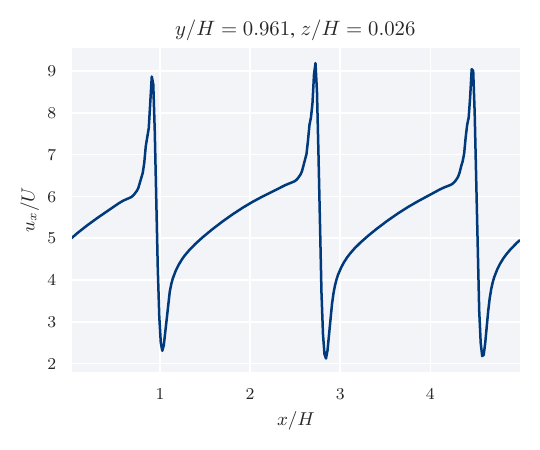}
    \caption{Jet detachments represented by the streamwise velocity $u_x/U$ over the duct length $x/H$ in a duct flow with  $\reyn = 2000$ and $\ha = 2000$.}
    \label{fig:app:val:duct_u_x_over_x}
\end{figure}

\subsubsection{Convection in a Tall Box}
Following the benchmark configuration in \citet{krasnov2023}, we consider natural convection of liquid metal (mercury) in a tall box with dimensions $L_x \times L_y \times L_z = 1 \times 7.5 \times 1$.
Gravity acts in the negative $y$-direction, while an external uniform magnetic field is applied along the horizontal $x$-axis.
Constant temperatures are applied at the vertical walls perpendicular to $z$ ($T = 0.5$ at the hot wall, $T = -0.5$ at the cold wall), with all other walls thermally and electrically insulated.
The flow is simulated at $\mathrm{Gr} = 3 \times 10^7$ ($\mathrm{Ra} = 7.5 \times 10^5$) across varying magnetic field strengths spanning $0 \le \mathrm{Ha} \le 796$.

\begin{figure}[H]
    \centering
    \includegraphics[width=\textwidth]{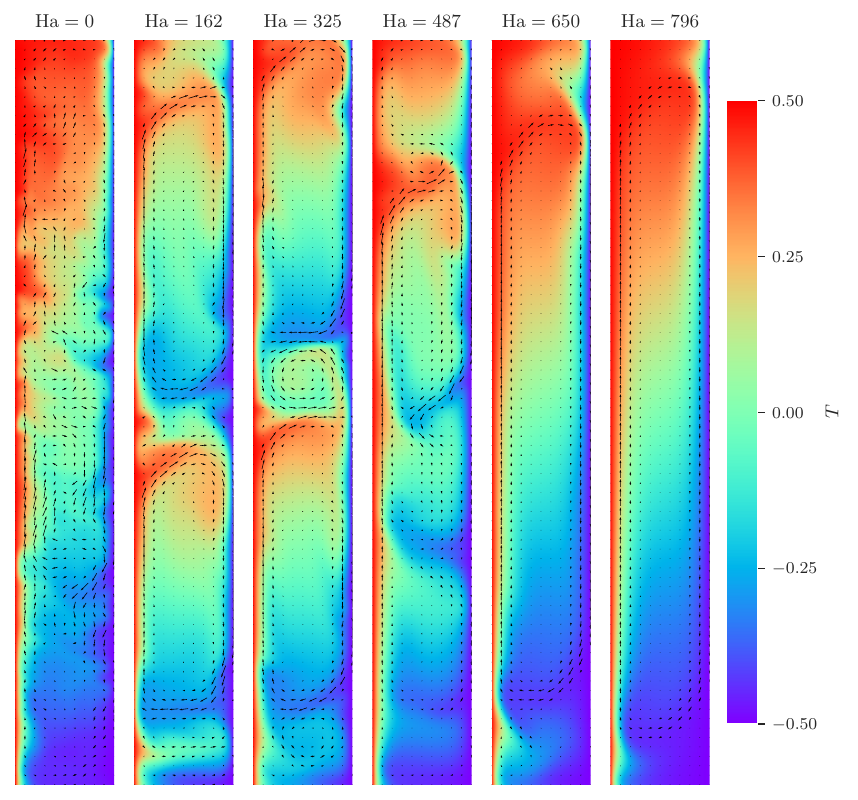}
    \caption{Convection case I: Temperature and velocity arrows for different Hartmann numbers.}
    \label{fig:app:val:temp1_temperature}
\end{figure}

As summarized in Figure~\ref{fig:app:val:temp1_temperature}, our computed Nusselt numbers ($\mathrm{Nu}$) show strong agreement with the reference values across most regimes ($\epsilon < 1.4\%$).
At intermediate values ($\mathrm{Ha} = 162$ and $\mathrm{Ha} = 487$), the flow is inherently unsteady and subject to transitions between multiple flow states \citep{krasnov2023}, leading to higher relative deviations.

\begin{table}[h]
    \centering
    \caption{Experimental results for the convection in a box case \citep{krasnov2023}. $^*$In these cases, the flow is unstable and does not converge to a steady-state solution, as pointed out by \citet{krasnov2023}. Therefore, our results do not match the reference numbers.}
    \label{tab:app:val_temp1}
    \input{tab/val_temp1}
\end{table}

\subsubsection{Convection in a Square Box}
The second configuration evaluates horizontally driven natural convection of a fluid in a unit cubic cavity ($L_x \times L_y \times L_z = 1 \times 1 \times 1$) at $\mathrm{Gr} = 3.15 \times 10^6$ ($\mathrm{Ra} = 10^5$) and $\mathrm{Ha} = 100$ \citep{krasnov2023}.
While maintaining the same orthogonal orientation of gravity ($-y$), magnetic field ($x$), and temperature gradient ($z$), this setup benchmarks the effect of finite wall electrical conductivity by varying the uniform wall conductance ratio $C_w \in [0, 50]$, alongside a hydrodynamic baseline (No MHD).

As shown in Table~\ref{tab:app:val_temp2}, our steady-state results closely match the reference data across all conductance ratios, with relative errors remaining below $0.84\%$.

\begin{figure}[H]
    \centering
    \includegraphics[width=\textwidth]{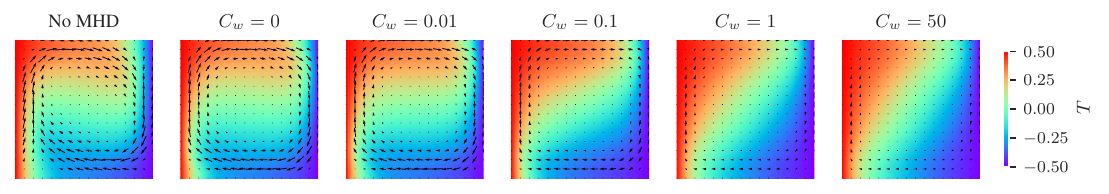}
     \caption{Convection case II: Temperature and velocity arrows for different wall conductance ratios.}
    \label{fig:app:val:temp2_temperature}
\end{figure}

\begin{table}[h]
    \centering
    \caption{Experimental results for the convection in a box case \citep{krasnov2023}.}
    \label{tab:app:val_temp2}
    \input{tab/val_temp2}
\end{table}

\subsection{Backward Validation}

\begin{figure}[H]
    \centering
    \includegraphics[width=\textwidth]{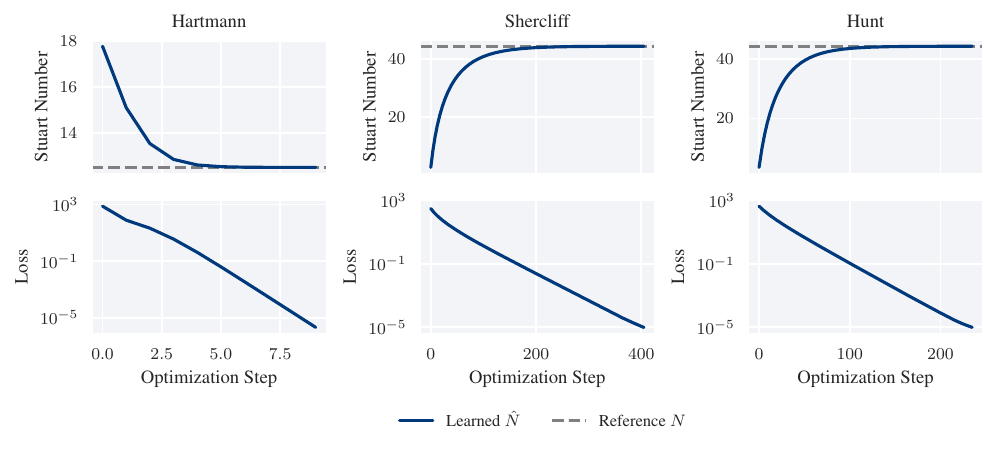}
    \caption{Stuart number $\stuart$ learning results for validating solver gradients. Gray dashed line indicates the Stuart number of the reference simulation.}
    \label{fig:app:val:optim_results}
\end{figure}

Similar to the PISO steps in PICT~\citep[cf.][]{franz2026}, we use the PyTorch \texttt{gradcheck} tool to validate the \solver{} gradient operations.
Additionally, we perform a learning task where we learn a physical parameter through the differentiable MHD simulation (see Figure~\ref{fig:app:val:optim_results}) on three previously discussed flow settings.
The goal is to learn the Stuart number $\stuart$ of a reference simulation given the final velocity field after a fixed number of steps, where the loss function is the MSE between the velocity of the optimized and the reference simulation.
Based on the loss with respect to the reference, a backward pass through the differentiable MHD simulation is performed, and the learned Stuart number $\hat N$ is updated using SGD.
The procedure is repeated until the loss is below $10^{-5}$.

%% file: tab/val_hartmann.tex
\begin{tabular}{c|cc|c|cc} 
    \toprule
    $\mathrm{\mathbf{Ha}}$ &
    $\mathbf{N_x}$ &
    $\mathbf{N_y}$ &
    $\mathbf{E_y}$ &
    $\MAEbf$ \\
    \midrule
    0.00 & 100 & 50 & 10 & 0.04 \\
    5.00 & 100 & 50 & 10 & 0.02 \\
    10.00 & 100 & 50 & 10 & 0.03 \\
    20.00 & 100 & 50 & 10 & 0.08 \\
    20.00 & 200 & 100 & 10 & 0.02 \\
    20.00 & 400 & 200 & 10 & 0.01 \\
    \bottomrule
\end{tabular}

%% file: tab/val_shercliff.tex
\begin{tabular}{c|ccc|cc|cc} 
    \toprule
    $\mathrm{\mathbf{Ha}}$ &
    $\mathbf{N_x}$ &
    $\mathbf{N_y}$ &
    $\mathbf{N_z}$ &
    $\mathbf{E_y}$ &
    $\mathbf{E_z}$ &
    $\MAEbf$ &
    $\RMSEbf$ \\
    \midrule
    20   & 50 & 40 & 40 & 200 & 50 & 0.22 & 0.48 \\
    100  & 50 & 40 & 60 & 200 & 50 & 0.31 & 0.65 \\
    1000 & 50 & 50 & 50 & 200 & 50 & 7.98 & 10.97 \\
    1000 & 50 & 100 & 100 & 200 & 50 & 2.07 & 2.75 \\
    1000 & 100 & 150 & 150 & 200 & 50 & 0.75 & 0.94 \\
    \bottomrule
\end{tabular}

%% file: tab/val_hunt.tex
\begin{tabular}{c|ccc|cc|cc} 
    \toprule
    $\mathrm{\mathbf{Ha}}$ &
    $\mathbf{N_x}$ &
    $\mathbf{N_y}$ &
    $\mathbf{N_z}$ &
    $\mathbf{E_y}$ &
    $\mathbf{E_z}$ &
    $\MAEbf$ &
    $\RMSEbf$ \\
    \midrule
    20   & 50 & 40 & 40 & 200 & 50 & 0.15 & 0.28 \\
    100  & 50 & 40 & 60 & 200 & 50 & 0.11 & 0.31 \\
    1000 & 50 & 50 & 50 & 200 & 50 & 0.11 & 0.94 \\
    1000 & 50 & 100 & 100 & 200 & 50 & 0.03 & 0.22 \\
    1000 & 100 & 150 & 150 & 200 & 50 & 0.02 & 0.10 \\
    \bottomrule
\end{tabular}

%% file: tab/val_temp1.tex
\begin{tabular}{c|c|c|c} 
    \toprule
    $\mathbf{Ha}$ & 
    $\mathbf{Nu}$ & 
    $\mathbf{Nu_{ref}}$
    & $\boldsymbol{\epsilon}\,$(\%) \\
    \midrule
    0   & 3.751 & 3.800 & 1.302 \\
    162 & 4.453 & 5.150 & 13.529$^*$ \\
    325 & 4.417 & 4.440 & 0.518 \\
    487 & 3.621 & 3.440 & 5.265$^*$ \\
    650 & 3.181 & 3.150 & 0.975 \\
    796 & 3.034 & 3.010 & 0.812 \\
    \bottomrule
\end{tabular}

%% file: tab/val_temp2.tex
\begin{tabular}{c|c|c|c} 
    \toprule
    $\mathbf{C_w}$ & 
    $\mathbf{Nu}$ & 
    $\mathbf{Nu_{ref}}$
    & $\boldsymbol{\epsilon}$\,(\%) \\
    \midrule
    No MHD        & 3.359 & 3.331 & 0.838 \\
    0             & 3.241 & 3.245 & 0.139 \\
    0.01          & 3.040 & 3.046 & 0.183 \\
    0.1           & 2.276 & 2.277 & 0.050 \\
    1             & 1.537 & 1.527 & 0.681 \\
    50 & 1.387 & 1.390 & 0.184 \\
    \bottomrule
\end{tabular}

%% file: content/app_experiments.tex
\subsection{Hardware and Software Configuration}
\paragraph{Reproducibility and Data Availability.}
All validation, RL, and inverse design results are publicly available in our HuggingFace dataset\footnote{\url{https://huggingface.co/datasets/upb-sciml/mhd_control}}.
Further, all scripts to run the experiments presented in this work are available on GitHub\footnote{\url{https://github.com/upb-sciml/fluidgym/tree/mhd_control/runscripts}}.

\paragraph{General Experimental Setup.}
Experiments were conducted on the following types of nodes with the following specifications:
A100-40GB nodes:
\begin{itemize}
    \item \textbf{Python:} 3.12.13
    \item \textbf{PyTorch:} 2.10.0+cu128
    \item \textbf{CUDA:} 12.8
    \item \textbf{System Memory:} 32\,GB RAM
    \item \textbf{CPU:} 32 cores of an AMD EPYC 7763
    \item \textbf{GPU:} 1 $\times$ NVIDIA A100 (40\,GB)
\end{itemize}
A100-80GB nodes:
\begin{itemize}
    \item \textbf{Python:} 3.12.13
    \item \textbf{PyTorch:} 2.10.0+cu128
    \item \textbf{CUDA:} 12.8
    \item \textbf{System Memory:} 32\,GB RAM
    \item \textbf{CPU:} 32 cores of an AMD EPYC 7742
    \item \textbf{GPU:} 1 $\times$ NVIDIA A100 (80\,GB)
\end{itemize}
All SAC runs were computed on A100-40GB nodes and BPTT and SHAC experiments were conducted on A100-80GB nodes.

\subsection{Control Algorithm Hyperparameters}
\label{sec:app:experiments:rl_hps}

\input{tab/sac_hps}
\input{tab/bptt_hps}
\input{tab/shac_hps}

\subsection{Inverse Design Optimizer Hyperparameters}
\label{sec:app:experiments:optim_hps}

\input{tab/cmaes_hps}
\input{tab/lbfgsb_hps}

%% file: tab/sac_hps.tex
\begin{table}[H]
  \caption{SAC hyperparameters used in all RBC-MHD experiments.}
  \label{tab:sac_hps}
  \begin{center}
    \begin{small}
      \begin{sc}
        \begin{tabular}{lc}
            \toprule
            \textbf{Hyperparameter} & \textbf{Value} \\
            \midrule
            Policy Network Type & \texttt{MlpPolicy} \\
            Policy Network Layers & \texttt{[256, 256]} \\
            Critic Network Layers & \texttt{[256, 256]} \\
            Learning Rate & $3 \times 10^{-4}$ \\
            Discount Factor ($\gamma$) & $0.99$ \\
            Soft Update Coefficient ($\tau$) & $0.005$ \\
            Replay Buffer Size & $5 \times 10^4$ \\
            Batch Size & $256$ \\
            Learning Starts & $100$ \\
            Training Frequency & $1$ \\
            Gradient Steps per Update & $-1$ (equal to \texttt{train\_freq}) \\
            Entropy Coefficient ($\alpha$) & \texttt{auto} \\
            Target Entropy & \texttt{auto} \\
            Target Update Interval & $1$ \\
            Device & \texttt{CUDA} \\
            \bottomrule
        \end{tabular}
      \end{sc}
    \end{small}
  \end{center}
  \vskip -0.1in
\end{table}

%% file: tab/bptt_hps.tex
\begin{table}[H]
  \caption{BPTT hyperparameters used in all RBC-MHD experiments.}
  \label{tab:bptt_hps}
  \begin{center}
    \begin{small}
      \begin{sc}
        \begin{tabular}{lc}
            \toprule
            \textbf{Hyperparameter} & \textbf{Value} \\
            \midrule
            Policy Network Layers & \texttt{[128, 64, 32]} \\
            Hidden Layer Normalization & \texttt{LayerNorm} \\
            Hidden Activation & \texttt{ELU} \\
            Output Activation & \texttt{Tanh} \\
            Optimizer & \texttt{AdamW} \\
            Learning Rate & $10^{-3}$ \\
            Discount Factor ($\gamma$) & $0.999$ \\
            Horizon & $10$ \\
            Reward Scale & $1.0$ \\
            Max. Gradient Norm & $0.5$ \\
            Observation Normalization & \texttt{True} \\
            Detach Observations & \texttt{True} \\
            Checkpoint Segment Size & $5$ \\
            Device & \texttt{CUDA} \\
            \bottomrule
        \end{tabular}
      \end{sc}
    \end{small}
  \end{center}
  \vskip -0.1in
\end{table}

%% file: tab/shac_hps.tex
\begin{table}[H]
  \caption{SHAC hyperparameters used in all RBC-MHD experiments.}
  \label{tab:shac_hps}
  \begin{center}
    \begin{small}
      \begin{sc}
        \begin{tabular}{lc}
            \toprule
            \textbf{Hyperparameter} & \textbf{Value} \\
            \midrule
            Actor Network Layers & \texttt{[256, 256]} \\
            Critic Network Layers & \texttt{[256, 256]} \\
            Hidden Layer Normalization & \texttt{LayerNorm} \\
            Hidden Activation & \texttt{ELU} \\
            Initial Log-Std. ($\log \sigma_0$) & $-2.3$ \\
            Tanh Action Clamping & \texttt{True} \\
            Observation Normalization & \texttt{True} \\
            Optimizer & \texttt{Adam} \\
            Actor Learning Rate & $3 \times 10^{-4}$ \\
            Critic Learning Rate & $3 \times 10^{-4}$ \\
            Learning Rate Schedule & \texttt{constant} \\
            Discount Factor ($\gamma$) & $0.99$ \\
            Horizon & $10$ \\
            Critic Method & TD($\lambda$) \\
            TD Parameter ($\lambda$) & $0.95$ \\
            Critic Iterations & $16$ \\
            Critic Mini-Batches & $1$ \\
            Target Critic Coefficient ($\alpha$) & $0.2$ \\
            Max. Gradient Norm & $1.0$ \\
            Reward Scale & $1.0$ \\
            Checkpoint Segment Size & $5$ \\
            Device & \texttt{CUDA} \\
            \bottomrule
        \end{tabular}
      \end{sc}
    \end{small}
  \end{center}
  \vskip -0.1in
\end{table}

%% file: tab/cmaes_hps.tex
\begin{table}[H]
  \caption{CMA-ES hyperparameters for the inverse $C_w$ problem. Unlisted options keep the defaults of the \texttt{pycma} library.}
  \label{tab:cmaes_hps}
  \begin{center}
    \begin{small}
      \begin{sc}
        \begin{tabular}{lc}
            \toprule
            \textbf{Hyperparameter} & \textbf{Value} \\
            \midrule
            Implementation & \texttt{pycma} \\
            Parametrization & $\log_{10} C_w$ per patch \\
            Initial Step Size ($\sigma_0$) & $0.5$ \\
            Population Size & $4 + \lfloor 3 \ln n \rfloor$ (default) \\
            Bounds on $\log_{10} C_w$ & $[-3, \log_{10} 50]$ \\
            Stopping Objective & $10^{-5}$ \\
            Wall-Time Limit & $1800\,\mathrm{s}$ \\
            Gradients & \texttt{None} \\
            \bottomrule
        \end{tabular}
      \end{sc}
    \end{small}
  \end{center}
  \vskip -0.1in
\end{table}

%% file: tab/lbfgsb_hps.tex
\begin{table}[H]
  \caption{L-BFGS-B hyperparameters for the inverse $C_w$ problem. Gradients are computed by the differentiable solver. Unlisted options keep the SciPy defaults.}
  \label{tab:lbfgsb_hps}
  \begin{center}
    \begin{small}
      \begin{sc}
        \begin{tabular}{lc}
            \toprule
            \textbf{Hyperparameter} & \textbf{Value} \\
            \midrule
            Implementation & \texttt{scipy.optimize} \\
            Parametrization & $\log_{10} C_w$ per patch \\
            Function Tolerance (\texttt{ftol}) & $10^{-12}$ \\
            Gradient Tolerance (\texttt{gtol}) & $10^{-8}$ \\
            Max. Iterations & $15000$ \\
            Max. Function Evaluations & $15000$ \\
            Bounds on $\log_{10} C_w$ & $[-3, \log_{10} 50]$ \\
            Stopping Objective & $10^{-5}$ \\
            Wall-Time Limit & $1800\,\mathrm{s}$ \\
            Gradients & Differentiable solver \\
            \bottomrule
        \end{tabular}
      \end{sc}
    \end{small}
  \end{center}
  \vskip -0.1in
\end{table}

%% file: content/app_results.tex
\subsection{Learning Closed-loop Control: Liquid Metal Rayleigh-Bénard Convection}
\label{sec:app:results:rl}

\begin{figure}[H]
    \centering
    \includegraphics[width=\textwidth]{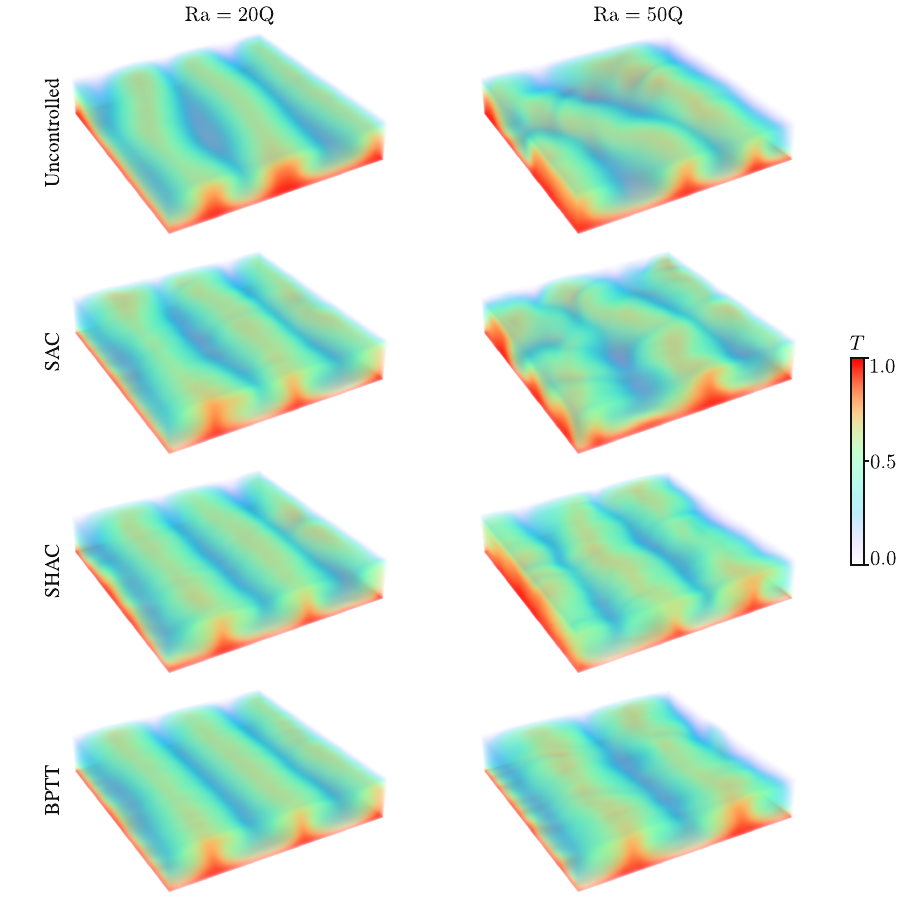}
    \caption{Qualitative visualizations of the final MHD RBC temperature fields controlled by the BPTT policy.}
    \label{fig:RBCMHD_renders_all}
\end{figure}

In addition to the overall test set reward comparison, we also consider the mean improvement in $\nusselt$ and the relative actuation power consumption.
These are denoted in Tables~\ref{tab:app:RBCMHD3D_test_nu_improvement} and~\ref{tab:app:RBCMHD3D_test_power_increase}, respectively.
Notably, while SAC achieves a greater heat transfer improvement, it also yields higher actuation power.
BPTT, on the contrary, achieves a much lower power consumption at lower Nusselt improvement, while still achieving the highest overall reward, which balances $\nusselt$ improvement and actuation power.

\begin{figure}[H]
    \centering
    \includegraphics[width=\textwidth]{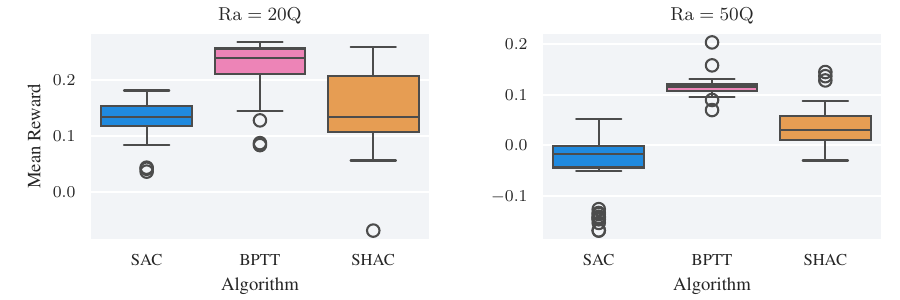}
    \caption{Per-step test set rewards for all algorithms across both regimes.}
    \label{fig:RBCMHD3D_test_reward}
\end{figure}

\begin{table}[H]
    \centering
    \caption{Interquartile mean relative improvement in $\nusselt$ over the uncontrolled baseflow over 10 test episodes.}
    \input{tab/test/RBCMHD3D_test_nu_improvement}
    \label{tab:app:RBCMHD3D_test_nu_improvement}
\end{table}

\begin{table}[H]
    \centering
    \caption{Interquartile mean relative actuation power increase over the uncontrolled baseflow over 10 test episodes. Lower is better.}
    \input{tab/test/RBCMHD3D_test_power_increase}
    \label{tab:app:RBCMHD3D_test_power_increase}

\end{table}

\subsection{Inverse Design: Detecting Defective Wall Patches in a Convection Box}
\label{sec:app:results:inverse_design}

\begin{figure}[H]
    \centering
    \includegraphics[width=\textwidth]{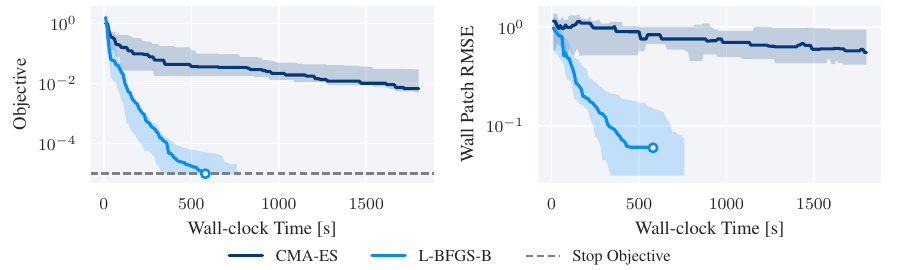}
    \caption{Defect detection optimization using CMA-ES and L-BFGS-B. \textbf{Left}: best objective $J$ over wall-clock time. \textbf{Right}: RMSE between true $C_w$ and optimizer $C_w$ map for best objective $J$. Lines and shading show median and min–max range across seeds, respectively.}
    \label{fig:app:results:optimization_run}
\end{figure}

%% file: tab/test/RBCMHD3D_test_nu_improvement.tex
\begin{tabular}{lccc}
\toprule
$\rayleigh$ & SAC & BPTT & SHAC \\
\midrule
$20\chandr$ & $\mathbf{+33.56}\%$ & $+28.59\%$ & $+31.60\%$ \\
$50\chandr$ & $+10.73\%$ & $+16.44\%$ & $\mathbf{+21.18}\%$ \\
\bottomrule
\end{tabular}

%% file: tab/test/RBCMHD3D_test_power_increase.tex
\begin{tabular}{lccc}
\toprule
$\rayleigh$ & SAC & BPTT & SHAC \\
\midrule
$20\chandr$ & $2015.72\%$ & $\mathbf{476.94}\%$ & $1558.45\%$ \\
$50\chandr$ & $1309.66\%$ & $\mathbf{477.40}\%$ & $1884.74\%$ \\
\bottomrule
\end{tabular}